\documentclass{article}

\usepackage[preprint]{neurips_2023}

\usepackage[utf8]{inputenc} 
\usepackage[T1]{fontenc}    
\usepackage{hyperref}       
\usepackage{url}            
\usepackage{booktabs}       
\usepackage{amsfonts}       
\usepackage{nicefrac}       
\usepackage{microtype}      
\usepackage{xcolor}         

\usepackage{multirow}
\usepackage{graphicx}
\usepackage{setspace}
\usepackage{algorithm}
\usepackage{algorithmic}
\usepackage{bbm}
\usepackage{caption}
\usepackage{subfigure}

\usepackage{amsmath,amsfonts,bm}

\def\eqref#1{equation~\ref{#1}}

\def\1{\bm{1}}

\def\vtheta{{\bm{\theta}}}

\def\vg{{\bm{g}}}

\def\vx{{\bm{x}}}
\def\vy{{\bm{y}}}
\def\vz{{\bm{z}}}

\DeclareMathAlphabet{\mathsfit}{\encodingdefault}{\sfdefault}{m}{sl}
\SetMathAlphabet{\mathsfit}{bold}{\encodingdefault}{\sfdefault}{bx}{n}

\def\gN{{\mathcal{N}}}

\newcommand{\E}{\mathbb{E}}
\newcommand{\Ls}{\mathcal{L}}

\usepackage[noabbrev,capitalize]{cleveref}

\crefname{equation}{equation}{equations}
\crefname{line}{line}{lines}
\crefname{section}{\S}{\S\S}
\Crefformat{section}{#2\S#1#3}
\crefrangeformat{section}{\S\S#3#1#4--#5#2#6}

\newcommand{\ours}{\textsc{AR-Diffusion}}

\title{\ours{}: Auto-Regressive Diffusion Model for Text Generation}

\author{
  Tong Wu$^{1}$\thanks{Work done during an internship at Microsoft Research Asia.}~~\thanks{These authors contributed equally to this work.}~~,
  Zhihao Fan$^{2*\dagger}$,
  Xiao Liu$^{3}$,
  Yeyun Gong$^{3}$\thanks{Corresponding author.}~,
  Yelong Shen$^{4}$,
  Jian Jiao$^{5}$, 
  \\ \textbf{
  Hai-Tao Zheng$^{1,8\ddagger}$
  Juntao Li$^{6}$,
  Zhongyu Wei$^{2}$,
  Jian Guo$^{7}$,  
  Nan Duan$^{3\ddagger}$,
  Weizhu Chen$^{4\ddagger}$  
  }
  \\
  $^1$Shezhen International Graduate School, Tsinghua University, $^2$ Fudan University, \\
  $^3$Microsoft Research Asia, $^4$Microsoft Azure AI, Redmond, $^5$Microsoft, \\
   $^6$Soochow University, $^7$IDEA Research
  \\
  \texttt{\{yegong, yeshe, nanduan, wzchen\}}@microsoft.com, \\
  \texttt{zheng.haitao}@sz.tsinghua.edu.cn,
}

\begin{document}

\maketitle

\begin{abstract}
Diffusion models have gained significant attention in the realm of image generation due to their exceptional performance. Their success has been recently expanded to text generation via generating all tokens within a sequence concurrently. 
However, natural language exhibits a far more pronounced sequential dependency in comparison to images, and the majority of existing language models are trained with a left-to-right auto-regressive approach.
To account for the inherent sequential characteristic of natural language, we introduce  Auto-Regressive Diffusion (\ours{}). \ours{} ensures that the generation of tokens on the right depends on the generated ones on the left, a mechanism achieved through employing a dynamic number of denoising steps that vary based on token position. This results in tokens on the left undergoing fewer denoising steps than those on the right, thereby enabling them to generate earlier and subsequently influence the generation of tokens on the right.
In a series of experiments on various text generation tasks, including text summarization, machine translation, and common sense generation, \ours{} clearly demonstrated its superiority over existing diffusion language models and that it can be $100\times\sim600\times$ faster when achieving comparable results. Our code is available at \url{https://github.com/microsoft/ProphetNet/tree/master/AR-diffusion}.
\end{abstract}

\section{Introduction}
Text generation is a fundamental task within the field of natural language processing (NLP). Pre-trained language models like GPT-4 \citep{DBLP:journals/corr/abs-2303-08774}, LLaMA \citep{DBLP:journals/corr/abs-2302-13971}, and Alpaca \citep{alpaca} have garnered significant attention with their ability to generate fluent and human-like textual content. These models utilize the auto-regressive (AR) Transformer decoders \citep{NIPS2017_3f5ee243} to emit generated tokens one-by-one in sequential order from left to right. By leveraging the power of position dependency, AR models are able to enhance the naturalness, coherence, and adherence to human language conventions in the generated text \citep{brown2020language}. 

Recent studies have shown the remarkable performance of diffusion models in image generation~\citep{DBLP:conf/nips/HoJA20}, motivating researchers to extend diffusion to text generation~ \citep{li2022diffusionlm,DBLP:journals/corr/abs-2210-08933,dieleman2022continuous,yuan2022seqdiffuseq,ye2023dinoiser}. By introducing timestep, these methods progressively regulate the interpolation between the original tokens and Gaussian noise, then iteratively denoise for text generation. At each timestep, the diffusion-based text generator predicts all tokens simultaneously following Non-Auto-Regression (NAR)~\citep{DBLP:conf/acl/LewisLGGMLSZ20,qi-etal-2020-prophetnet,qi2021bang,DBLP:conf/emnlp/LiTZNW22}, leading to faster decoding speed compared to AR. However, it also inherits the drawback of NAR, namely the sacrifice of inter-token position dependency~\citep{DBLP:conf/emnlp/LiCY022} and the drop of generation performance \citep{bao-etal-2021-non}. 

\begin{figure}[t]
\centering
\includegraphics[width=\linewidth]{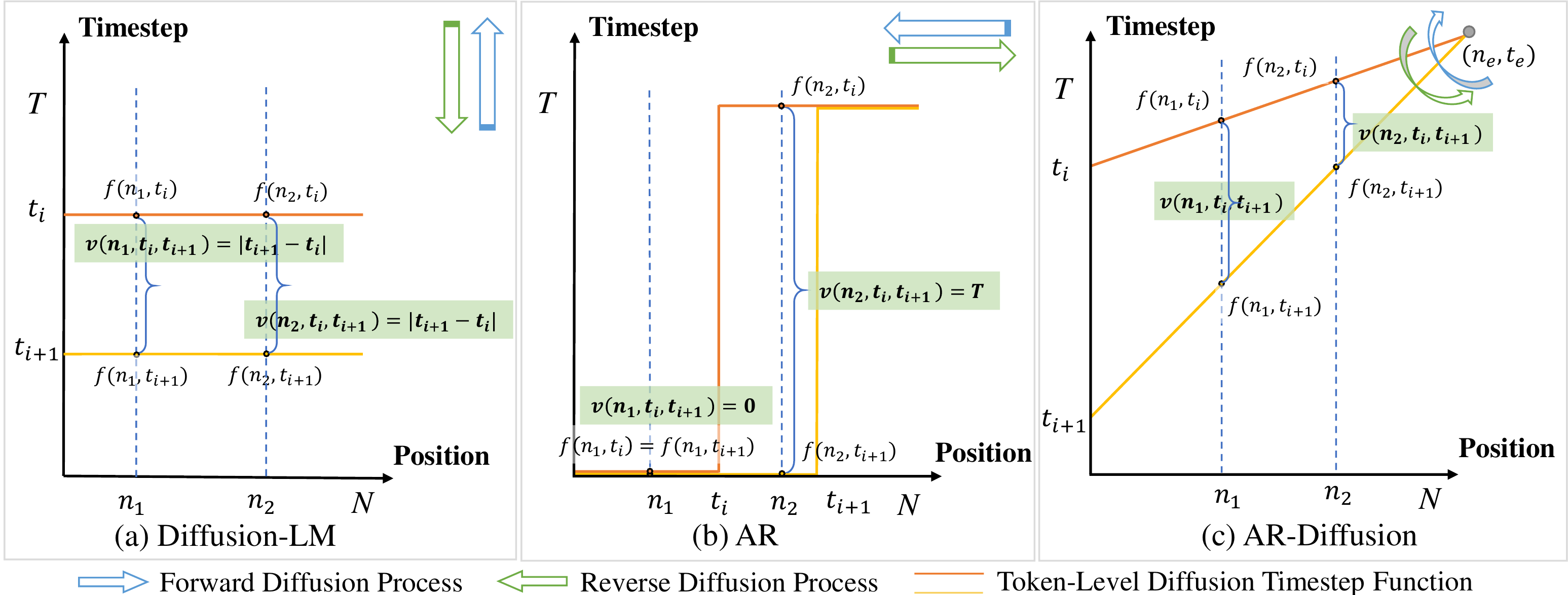}
\caption{Model behaviors illustrated on a two-dimensional coordinate system, where the horizontal axis stands for the position and the vertical axis represents the diffusion timestep.
In the inference stage, different models will behave differently.
(a) For the typical Diffusion-LM \citep{li2022diffusionlm}, each token share the identical movement speed $v(n_1,t_{i},t_{i+1})=v(n_2,t_{i},t_{i+1})=|t_{i+1}-t_i|$.
(b) For AR from the perspective of diffusion models, the tokens have two states based on the degree of interpolation between the original tokens and Gaussian noise: to be decoded (at timestep $t=T$) and already decoded (at timestep $t=0$). Specifically, we have $v(n_1,t_{i},t_{i+1})=0$ and $v(n_2,t_{i},t_{i+1})=T$.
(c) In \ours, $(n_e, t_e)$ is the coordinate of anchor point. Tokens in different positions exhibit varying movement speeds, such as $v(n_1,t_{i},t_{i+1}) > v(n_2,t_{i},t_{i+1})$ when $n_1 < n_2$. \looseness=-1
}
\label{framework}
\end{figure}

To conduct a comprehensive analysis, we introduce a two-dimensional coordinate system to track the diffusion timestep of tokens $f(\cdot)$ positioned at various locations. As illustrated in~\cref{framework}, the system assigns the token position $n \in \left[1, N\right]$ to the horizontal axis and the diffusion timestep $t \in [0, T]$ to the vertical axis. Diffusion-LM \citep{li2022diffusionlm}, which is followed by existing diffusion-based text generation models, is shown in~\cref{framework}(a). It assigns a uniform timestep $t$ to all tokens. In contrast, tokens in the AR model depicted in~\cref{framework}(b) exhibit distinct timesteps within a generation step ($t_{i}$). For instance, the already decoded token at position $n_1$ has a timestep of $0$, while the to-be-decoded token at position $n_2$ has a timestep of $T$. This approach effectively captures the sequential dependency. Motivated by this observation, we introduce \ours{}, an auto-regressive diffusion method, for the disparity in token positions and the principle of sequential token identification.

In \ours{}, we propose a \textbf{multi-level diffusion strategy} that includes both sentence-level and token-level diffusion. We randomly choose a sentence-level timestep $t$, and assign \textbf{dynamic movement speeds} $v(\cdot)$ by determining position-sensitive token-level timestep $f(n,t)$ for each token. This enables tokens at the left of a sentence to undergo faster movement from random Gaussian noise to token embedding, while those at the right of the sentence experience slower movement to better utilize information from previously denoised tokens. During inference, to reduce the significant number of inference steps (e.g., 2,000) required in Diffusion-LM \citep{li2022diffusionlm}, SeqDiffSeq \citep{yuan2022seqdiffuseq} and GENIE~\citep{lin2023text}, we introduce a skipping mechanism that collaborates with the multi-level diffusion strategy to accelerate the process.

Experimental results across various text generation tasks, such as text summarization, machine translation, and common sense generation, have consistently demonstrated that \ours{} surpasses existing text diffusion models, including AR methods in terms of both quality and diversity. Moreover, our verification reveals that \ours{} requires fewer resources during decoding while maintaining superior performance. It achieves $100\times$ faster than SeqDiffSeq \citep{yuan2022seqdiffuseq} in machine translation and $600\times$ faster than GENIE \citep{lin2023text} in text summarization while delivering comparable results. Furthermore, it demonstrates promising results even in a challenging scenario where decoding is limited to only two steps. \looseness=-1

\section{Preliminary}
\subsection{Conditional Generative Language Models}
In the field of natural language generation, conditional generative models are commonly implemented using either auto-regressive (AR) or non-auto-regressive (NAR) methods. In AR \citep{NIPS2017_3f5ee243}, tokens on the right are predicted based on visible left tokens. The likelihood is given by $p_{\text{AR}}(\vy|\vx)=\prod_{i=1}^{N}p(\vy_i|\vy_{1:i-1}; \vx)$, where $y_i$ denotes the $i$-th token of $\vy$. On the other hand, NAR \citep{gu2017non} assumes conditional independence among tokens and generates them uniformly without distinction during decoding, resulting in the likelihood $p_\text{NAR}(\vy|\vx)=\prod_{i=1}^{N}p(\vy_i|\vx)$. This parallel generation approach is of lower quality compared to AR, although it offers a substantial speed advantage. \looseness=-1

\subsection{Diffusion Models for Text Generation}
Recently, \cite{li2022diffusionlm} propose a natural language generation model based on the diffusion process, which is typically divided into a forward noising process and 
a reverse denoising process. 

Specifically, the forward process is a fixed linear Gaussian model, which gradually perturbs the random variable $\vz_0$ until it becomes the standard Gaussian distribution. This can be formalized as:
\begin{equation}
\small
q(\vz_t \mid \vz_0; \vx) = \gN(\vz_{t};\sqrt{\bar{\alpha}_t} \vz_0, (1-\bar{\alpha}_t) \textbf{I}),
\label{equ:forwarddiff}
\end{equation}
where, $\bar{\alpha}_{t}=\prod_{i=1}^t\alpha_i$, and $\alpha_i$ is a coefficient that monotonically decreases with timestep $t$, $\vz_t$ is the  latent state at timestep $t$.

The reverse process is to initiate from standard Gaussian noise and progressively utilize the denoising transition $p_\vtheta(\vz_{t-1}|\vz_t;\vx)$ for generation.
\begin{equation}
\small
p_\vtheta(\vz_{t-1}\mid\vz_t;\vx) = \mathcal{N}\big(\vz_{t-1};\mu_\vtheta(\vz_t,t;\vx),\Sigma_\vtheta(\vz_t,t;\vx)\big), 
\label{equ:reversediff}
\end{equation}
where the mean $\mu_\vtheta$ and variance $\Sigma_\vtheta$ are learned from the model. In particular, we follow \cite{li2022diffusionlm}'s approach of using predefined variance without trainable parameters.

To extend the continuous diffusion process to discrete text generation,~\citet{li2022diffusionlm} introduce an additional Markov transition from the discrete tokens $\vy$ to the latent variable $\vz_0$. 
In practice, we add an embedding step $q_{\phi}(\vz_0|\vy)=\mathcal{N}(\vz_0;\mathrm{Emb}(\vy),(1-\alpha_0)\textbf{I})$ in the forward process, and use a trainable rounding step which is parametrized by $p_{\vtheta}(\vy|\vz_0;\vx)=\prod_{i=1}^{N} p_{\vtheta}(y_i| z_0^i;\vx)$ in the reverse process.
In each timestep, we utilize an encoder-decoder model $\vg_{\vtheta}(\vz_t,t;\vx)$ to approximate $\vz_0$ \citep{lin2023text} in a NAR manner and then estimate $\mu_\vtheta(\vz_t,t;\vx)$. 

In consequence, combined with maximizing the evidence lower bound (ELBO) of $\log p_{\vtheta}(\vy|\vx)$, our training objective of the conditional diffusion language model is:
\begin{equation}
\small
\Ls = \E_{q_\phi(\vz_{0:T} \mid \vy)} \left[ -\log p_{\vtheta}(\vy \mid \vz_0;\vx) + \sum_{t=1}^T \Vert \vz_0-\vg_\theta(\vz_t,t;\vx) \Vert^2 \right].
\label{equ:difflmloss}
\end{equation}

\section{Methodology}
\subsection{Multi-Level Diffusion}
In the typical diffusion process, every token in the text sequence has the same diffusion timestep. In order to leverage the sequential nature of language, we enable tokens to have different diffusion timesteps during the forward and reverse pass. To accomplish this, we propose a multi-level diffusion strategy that includes both sentence-level and token-level diffusion.

Firstly, at the sentence-level, we follow Diffusion-LM~\citep{li2022diffusionlm} to randomly select a timestep $t$.
Secondly, at the token-level, we incorporate positional information $n \in \left[1, N\right]$ based on the sentence-level timestep to regulate the diffusion timestep for the current token.
The procedure is illustrated as: \looseness=-1
\begin{equation}
\small
\vz_t=\big(\vz^{1}_{f(1,t)}, \vz^{2}_{f(2,t)},\cdots, \vz^{N}_{f(N,t)}\big),
\label{equ:reconddiff}
\end{equation}
where
$N$ is the given target sentence length,
$\vz_t$ is the sentence representation at timestep\footnote{Please note that if we talk about a ``timestep'' without explicitly indicating that it is for token-level, it should be for sentence-level.} $t$, 
$\vz_{f(n,t)}^n$ is the latent representation for the $n$-th token at sentence-level timestep $t$,
and $f(n, t)$ is a token-level timestep function that denotes the token-level diffusion timestep determined by token position $n$ and sentence-level timestep $t$.

We visualize the token-level timestep $\big(n, f(n,t)\big)$ onto a two-dimensional coordinate system as~\cref{framework} , which takes the token \textbf{position} as the horizontal axis and the sentence-level \textbf{timestep} as the vertical axis.
Furthermore, to provide a more profound description of the characteristics of movement, we define the speed of movement as the following equation.
\begin{equation}
\small
v(n, t_i, t_{i+1})=f(n,t_{i+1})-f(n,t_{i}),
\label{equ:speed}
\end{equation}
where $t_i$ and $t_{i+1}$ are the start and end sentence-level diffusion timesteps. It can be observed that tokens in Diffusion-LM share the same movement speed, while those in AR exhibit different speeds.\looseness=-1

\subsection{Token-Level Diffusion with Dynamic Movement Speed}

Based on the speed of movement, we propose a fundamental principle, dynamic movement speed, for designing the token-level diffusion timestep function $f(n,t)$ to take advantage of AR in diffusion. Specifically, elements on the left side of a sentence undergo higher movement speed from random Gaussian noise to token embedding, while those on the right side experience lower movement speed, thereby they can be generated in the later sentence-level timestep and utilize information from previously generated tokens more effectively.  \looseness=-1

\begin{algorithm}[!ht]
\small
\caption{Training Process of \ours{}.}
\label{alg:algorithm}
\hspace*{\algorithmicindent} 
\textbf{Input}: Dataset $\{(\vx, \vy)\}$, maximum timestep number $T$ and maximum target length $N$. \\
\hspace*{\algorithmicindent} 
\textbf{Output}: Optimized model parameters $\vtheta$.

\begin{algorithmic}[1]
\STATE Define an anchor point $(n_e, t_e)$\protect\footnotemark.
\REPEAT
\STATE Sample ($\vx, \vy$) from the dataset and embed $\vy$ into $\vz_0$.
\STATE Sample a sentence-level timestep $t$ from the interval $[0, N+T]$, then the start point is determined by the following equation: 
\begin{equation}
\small
(n_s, t_s)=\big(\mathrm{clip}(N-t, 0, N), \mathrm{clip}(t-N, 0, T)\big)
\label{equ_point_slope_start_point}
\end{equation}
\STATE Use the point-slope linear function to determine the token-level timestep $f(n, t)$ in position $n$:
\begin{equation}
\small
f(n, t)=\mathrm{clip}\big(\frac{t_e-t_s}{n_e-n_s}(n-n_s)+t_s, 0, T\big)
\label{equ_point_slope}
\end{equation}
\STATE Sample $\vz_{f(n,t)}^n$ for each $n$ in different positions with Gaussian reparameterization.
\STATE According to \cref{equ:difflmloss} and \cref{equ:decomposition}, employ gradient descent to optimize the objective: 
\begin{equation}
\small
\min\limits_\theta\Big[-\log p_\vtheta(\vy\mid \vz_{0}; \vx)+ \sum_{n=1}^{N} \big\|\vg_{\vtheta}(\vz^n_{f(n,t)}, f(n,t);\vx)- \vz_0\big\|^2\Big]
\label{equ:gradient}
\end{equation}
\UNTIL{converged}
\end{algorithmic}
\label{algorithm:1}
\end{algorithm}

\footnotetext{In particular, the anchor point is set as $(2\times N, T)$ in our implementation.}

Following the guidance of the principle, we develop a token-level diffusion strategy with the linear function, which is shown in \cref{framework}(c). In particular, the procedure is illustrated in \cref{algorithm:1}, where $\mathrm{clip}(x, \mathrm{min}, \mathrm{max})$ function is to clamp all elements in $x$ into the range $[ \mathrm{min}, \mathrm{max} ]$. Specifically, in the forward process of diffusion, the start point goes to the left from $(N, 0)$ to $(0, 0)$ along the horizontal axis and then moves up to $(0, T)$ along the vertical axis. Therefore, the entire range of sentence-level timestep is extended to $\left[0, N+T\right]$. \looseness=-1

In the reverse diffusion process, the multi-level diffusion follows the formula:
\begin{equation}
\small
\vg_{\vtheta}\big(\vz_{t},t;\vx\big)=\vg_{\vtheta}\Big(\big(\vz^1_{f(1,t)}, f(1,t)\big), \big(\vz^2_{f(2,t)}, f(2,t)\big), \cdots, \big(\vz^N_{f(N,t)}, f(N,t)\big);  \vx\Big),    
\label{equ:decomposition}
\end{equation}
where $\vg_{\theta}(\vz^n_{f(n,t)}, f(n,t);\vx)$ denotes the $n$-th element.

\subsection{Inference with Skipping}

Typically, the generation process needs to go through all the sentence-level timesteps from $T+N$ to $0$. To reduce the decoding time, we introduce a skipping mechanism that allows us to traverse a subset of timesteps. \looseness=-1

To ensure consistency between training and inference, we also need to calculate the timestep for each token during the inference process. Therefore, we first establish an anchor point, and then uniformly select a decreasing subsequence $\{t_i\}^M_{i=0}$ from all timesteps ($T+N$ to $0$). The count of this sequence is the total decoding steps $M$ ($M \ll T+N$). For example, assuming the interval is $500$ and $T+N$ is $2500$, then $M$ is $5$, and the subsequence is $[2500, 2000, 1500, 1000, 500, 0]$.

Each element of this subsequence represents the sentence-level timesteps $t$, and we can use \cref{equ_point_slope_start_point} to calculate $(n_s, t_s)$. Then, based on \cref{equ_point_slope}, we calculate the token-level timesteps corresponding to each position. We take the current sentence-level timestep $t_i$ and the next sentence-level timestep $t_{i+1}$, and calculate the token-level timesteps $f(n, t_i)$ and $f(n, t_{i+1})$ for each position. Since $M \ll T+N$, $t_{i+1} \ll t_i$, implying that $f(n, t_i) \ll f(n, t_{i+1})$. The essence of Skipping is reflected in the fact that each token undergoes significant span during denoising (from $f(n, t_i)$ to $f(n, t_{i+1})$).

\begin{algorithm}[!ht]
\small
\caption{Inference Process of \ours{} with the Skipping Mechanism.}
\label{alg:algorithm2}
\hspace*{\algorithmicindent} 
\textbf{Input}: Source condition $\vx$, number of decoding steps $M$ and model parameters $\vtheta$. \\
\hspace*{\algorithmicindent} 
\textbf{Output}: Predicted target embedding $\hat{\vy}$.

\begin{algorithmic}[1]
\STATE Define an anchor point $(n_e, t_e)$.
\STATE Uniformly select a decreasing subsequence of timesteps $\{t_i\}^M_{i=0}$ ranging from $T+N$ to $0$, where $M\ll T+N$.
\STATE Sample $\vz_{t_0} \sim \mathcal{N}(\textbf{0}, \textbf{I})$.
\FOR{$i = 0$ to $M-1$}
\STATE Calculate the start point $(n_s, t_s)$ using \cref{equ_point_slope_start_point}.
\STATE Based on the current sentence-level inference steps $t_i$ and the next one $t_{i+1}$, assign token-level timesteps $f(n, t_i)$ and $f(n, t_{i+1})$ to token in position $n$ using \cref{equ_point_slope}.
\STATE Reverse sample $\vz_{t_{i+1}}=\big(\vz^{1}_{f(1,t_{i+1})}, \vz^{2}_{f(2,t_{i+1})},\cdots, \vz^{N}_{f(N,t_{i+1})}\big)$ from $p_{\theta}(\vz_{t_{i+1}}\mid \vz_{t_{i}};\vx)$ with the following formulas:
\begin{equation}
\small
\begin{aligned}
p_{\theta}(\vz_{t_{i+1}}\mid \vz_{t_{i}};\vx) &= \prod_{n=1}^{N} p_{\theta}\big(\vz^{n}_{f(n, t_{i+1})}\mid \vz^{n}_{f(n, t_{i})}; \vx\big)
\end{aligned}
\label{equ:reverse_sample2_1} 
\end{equation}
\begin{equation}
\small
\begin{gathered}
p_{\theta}\big(\vz^{n}_{f(n, t_{i+1})}\mid \vz^{n}_{f(n, t_{i})};\vx\big) \sim \mathcal{N}\big(\vz^{n}_{f(n, t_{i+1})};\lambda \vz_{f(n,t_{i})}^{n}+\mu\vg_{\theta}(\vz^n_{f(n,t)}, f(n,t);\vx), \sigma \textbf{I}\big) 
\end{gathered}
\label{equ:reverse_sample2_2}
\end{equation}
\ENDFOR
\STATE Map $\vz_{t_M}$ to the nearest embedding $\hat{\vy}$.
\end{algorithmic}
\label{algorithm:2}
\end{algorithm}

In practice, we propose an algorithm for the inference, illustrated in \cref{algorithm:2}. 
\begin{equation}
\small
\begin{gathered}
\lambda =\frac{\sqrt{\frac{\bar{\alpha}_{f(n,t_{i})}}{\bar{\alpha}_{f(n,t_{i+1})}}}(1-\bar{\alpha}_{f(n,t_{i+1})})}{1-\bar{\alpha}_{f(n,t_{i})}},~ \mu =\frac{\sqrt{\bar{\alpha}_{f(n,t_{i+1})}}(1-\frac{\bar{\alpha}_{f(n,t_{i})}}{\bar{\alpha}_{f(n,t_{i+1})}})}{1-\bar{\alpha}_{f(n,t_{i})}},~
\sigma = \frac{(1-\alpha_{f(n,t_{i})})(1-\bar{\alpha}_{f(n,t_{i+1})})}{1-\bar{\alpha}_{f(n,t_{i})}}\label{equ:sigma}
\end{gathered}
\end{equation}

In \cref{equ:reverse_sample2_1}, the conditional distribution of $\vz_{t_{i+1}}$ is inferred by $p_{\theta}(\vz_{t_{i+1}}|\vz_{t_{i}};\vx)$, and then we decompose it by positions due to the independent forward process of elements at different positions. From \cref{equ:reverse_sample2_2} to \cref{equ:sigma}, we establish the relationship between tokens at different timesteps, and the detailed derivation can be found in \cref{appendix:proof}. \looseness=-1

\section{Experiments}
\subsection{Tasks and Datasets}

\paragraph{Text Summarization}
This task involves taking a long document as input and generating a concise sentence as output. This requires models with the ability to identify important content and rewrite it in a condensed form. In our experiments, we use the publicly available \textsc{XSum} \citep{narayan-etal-2018-dont} and \textsc{Cnn/DailyMail} \cite{NIPS2015_afdec700} on GLGE\footnote{\url{https://microsoft.github.io/glge/}}, which is also named as GLGE-Easy.

\paragraph{Machine Translation}
Translation is a widely used sequence-to-sequence task. The input is a sequence of words in the source language, and the output is a sequence of corresponding words in the target language.
We choose the IWSLT 2014 dataset and the data processing method is to follow the scripts provided by fairseq\footnote{https://github.com/facebookresearch/fairseq/tree/main/examples/translation}. 

\paragraph{Common Sense Generation} In this task, the model is provided with a concept set consisting of objects and actions as input. The objective is to generate a sentence that incorporates these concepts and describes a realistic scenario. We use \textsc{CommonGen}\footnote{\url{https://inklab.usc.edu/CommonGen/}} dataset for evaluation.

\subsection{Experimental Details}
\paragraph{Model Setup}
Our model configuration is implemented based on \texttt{Transformer-base} \citep{NIPS2017_3f5ee243}. In particular, For
\textsc{XSum} and \textsc{Cnn/DailyMail}, we set the diffusion embedding dimension to 128. For IWSLT14, we use 64-dimensional diffusion embedding, 4 attention heads and 1024-dimensional feed-forward layers. For \textsc{CommonGen}, we adopt 64-dimensional diffusion embedding, 8 attention heads and 512-dimensional feed-forward layers.

\paragraph{Training and Inference}
In the training phase, we employ a square-root noise schedule and 2,000 diffusion steps \citep{li2022diffusionlm}. Specially, we use the tokenizer and vocabulary constructed by Byte Pair Encoding (BPE)\footnote{We train bpe on the training set, and follow the vocabulary size of \texttt{fairseq}, IWSLT14 is set to 10,000
.} \citep{kudo-richardson-2018-sentencepiece} for translation tasks. For other tasks, we adopt the tokenizer and vocabulary of \texttt{bert-base-uncased}. 

\paragraph{Baselines}
We set four groups of baselines: 

$\bullet$ NAR: NAT \citep{gu2017non}, iNAT \citep{lee2018deterministic}, CMLM \citep{ghazvininejad-etal-2019-mask}, LevT \citep{gu2019levenshtein} and CNAT \citep{bao-etal-2021-non};

$\bullet$ Semi-NAR: InsT \citep{stern2019insertion}, iNAT \citep{lee2018deterministic}, CMLM \citep{ghazvininejad-etal-2019-mask} and LevT \citep{gu2019levenshtein};

$\bullet$ AR: bRNN \citep{GuLLL16}, LSTM \citep{GreffSKSS17} and Transformer \citep{NIPS2017_3f5ee243};

$\bullet$ Diffusion: DiffusionLM \citep{li2022diffusionlm}, CDCD \citep{dieleman2022continuous}, SeqDiffuSeq \citep{yuan2022seqdiffuseq}, DINOISER \citep{ye2023dinoiser} and GENIE \citep{lin2023text}. 

\paragraph{Metrics}
We follow the approach of \cite{qi-etal-2020-prophetnet}\footnote{\url{https://github.com/microsoft/ProphetNet/tree/master/GLGE_baselines}} to evaluate the \textbf{ROUGE-1/2/L} of the summarization task. For the evaluation of translation tasks, we adopt the setting of SeqDiffuSeq \citep{yuan2022seqdiffuseq} to report BLEU score. In addition, we also calculate the SacreBLEU score according to the setting of DINOISER \citep{ye2023dinoiser} for comparison. For \textsc{CommonGen}, we employ \texttt{ROUGE-2/L}, \texttt{BLEU-3/4}, \texttt{METEOR}
and \texttt{SPICE} under the evaluation methods of \cite{lin-etal-2020-commongen}\footnote{\url{https://github.com/INK-USC/CommonGen/tree/master/evaluation/Traditional/eval_metrics}}. \looseness=-1

\paragraph{Training Parameters}
Our training parameters on different datasets are shown in \cref{train para table}. Our linear schedule warm up steps is 4,000 $\times N_{gc}$ , where $N_{gc}$ denotes gradient accumulation number. In addition, we use the AdamW (weight decay = 0.0) optimizer and dropout is 0.2. All experiments are implemented on 8 Tesla V100-32G. It takes about 20 hours to train \textsc{XSum} and \textsc{Cnn/DailyMail}, about 5 hours to train IWSLT14, and about 2 hours to train \textsc{CommenGen}.

\begin{table}[ht]
\centering
\caption{Training Parameter Settings. Batch Size = mini batch size $\times$ $ N_{gc}$ $\times$ GPU number, Optimized Steps = total steps / $N_{gc}$, and $N_{gc}$ is gradient accumulation number.}
\begin{tabular}{l|cccccc}
\toprule
Dataset & Lr \& Schedule & Batch Size & Optimized Steps & Target Length  \\ \midrule
\textsc{XSum}      & 8e-4 \& Cosine & 128$\times$3$\times$8 & 80,000 / 3 & 50  \\
\textsc{Cnn/DailyMail}  & 8e-4 \& Cosine & 80$\times$5$\times$8 & 100,000 / 5 & 180  \\
IWSLT14 \textsc{De}→\textsc{En}   & 2e-3 \& Cosine & 192$\times$2$\times$8 & 160,000 / 2 & 90  \\
IWSLT14 \textsc{En}→\textsc{De}   & 1.8e-3 \& Cosine & 768$\times$1$\times$8 & 60,000 & 90  \\
\textsc{CommonGen} & 3e-4 \& Constant & 384$\times$1$\times$8 & 40,000 & 54  \\ \bottomrule
\end{tabular}
\label{train para table}
\end{table}

\subsection{Main Results}
The results on different datasets are shown in \cref{xsum main table},  \cref{cnndm main table}, \cref{seqdiffuseq main table} and \cref{commongen main table}. The best result is \textbf{bolded} and the second-best result is \underline{underlined }. ``$k$'' indicates the number of generated candidate samples. It can be seen from the results in each table that \ours{} achieves the best performance. 

During the inference process, we utilize \textbf{20} inference steps and employ Minimum Bayes Risk (MBR) \citep{kumar2004minimum} decoding to select the best sample, following~\citep{li2022diffusionlm}. We choose MBR instead of the selection approach in GENIE, as GENIE picks up the best sample by calculating the maximum score for each generated one using ground truth, which introduces unfairness. To ensure a fair comparison, we re-implement GENIE using our configuration and perform inference with 20 steps. 

\begin{table}[ht]
\small 
\centering
\setlength\tabcolsep{12pt} 
\caption{Results on \textsc{XSum} test set. The results of NAR and Semi-NAR are from \cite{qi2021bang}, and the results of AR are from GLGE \citep{liu2021glge}.}
\begin{tabular}{l|c|cccc}
\toprule
\textbf{Methods} & \textbf{Pattern}  & \textbf{\texttt{ROUGE-1}} & \textbf{\texttt{ROUGE-2}} & \textbf{\texttt{ROUGE-L}} \\ \midrule                                    
NAT~\citep{gu2017non} &\multirow{4}{*}{NAR} &  24.0 & 3.9 & 20.3  \\ 
 iNAT~\citep{lee2018deterministic} & & 24.0 & 4.0 & 20.4  \\ 
 CMLM~\citep{ghazvininejad-etal-2019-mask} &  &  23.8 & 3.6 & 20.2  \\
 LevT~\citep{gu2019levenshtein}  & & 24.8 &  4.2 & 20.9 \\\midrule
InsT~\citep{stern2019insertion} & \multirow{4}{*}{Semi-NAR}&  17.7 & 5.2 & 16.1  \\ 
iNAT~\citep{lee2018deterministic}  & & 27.0 & 6.9 & 22.4  \\ 
CMLM~\citep{ghazvininejad-etal-2019-mask} & & 29.1 & 7.7 & 23.0  \\
LevT~\citep{gu2019levenshtein}  & & 25.3 & 7.4 & 21.5  \\\midrule
LSTM~\citep{GreffSKSS17} &\multirow{2}{*}{AR\protect\footnotemark} & 25.1 & 6.9 & 19.9 \\ 
Transformer~\citep{NIPS2017_3f5ee243} & &  30.5 & \underline{10.4} & 24.2   \\ 
 \midrule
GENIE \citep{lin2023text} ($k$ = 50) &  \multirow{3}{*}{Diffusion} & 29.3 & 8.3  & 21.9  \\ 
 \ours{} ($k$ = 50) & & \underline{31.7} & 10.1  & \underline{24.7}  \\
 \ours{} ($k$ = 500) & & \textbf{32.2} & \textbf{10.6}  & \textbf{25.2}  \\ \bottomrule
\end{tabular}
\label{xsum main table}
\end{table}
\footnotetext{Notably, although AR's beam search has a small beam, the search space may be larger than 50 or even 500.}

\begin{table}[ht]
\small 
\centering
\setlength\tabcolsep{13pt} 
\caption{Results on \textsc{Cnn/DailyMail} test set. The results of AR are from GLGE \cite{liu2021glge}.}
\begin{tabular}{l|c|cccc}
\toprule
\textbf{Methods} & \textbf{Pattern}  & \textbf{\texttt{ROUGE-1}} & \textbf{\texttt{ROUGE-2}} & \textbf{\texttt{ROUGE-L}}  \\ \midrule 
LSTM \citep{GreffSKSS17} &\multirow{2}{*}{AR} & 37.3 & 15.7 & 34.4  \\ 
Transformer \citep{NIPS2017_3f5ee243} & &  39.5 & \underline{16.7} & 36.7  \\ 
 \midrule
GENIE \citep{lin2023text} ($k$ = 50) &  \multirow{3}{*}{Diffusion} & 34.4 & 12.8  & 32.1  \\ 
\ours{} ($k$ = 50) & & \underline{39.6} & 16.3  & \underline{37.1} \\
\ours{} ($k$ = 500) & & \textbf{40.2} & \textbf{17.1}  & \textbf{37.7}  \\ \bottomrule
\end{tabular}
\label{cnndm main table}
\end{table}

\paragraph{Text Summarization} 
The results presented in \cref{xsum main table} and \cref{cnndm main table} clearly demonstrate that \ours{} outperforms the existing NAR and Semi-NAR approaches across all metrics. Moreover, \ours{} consistently achieves significant improvements over GENIE in terms of all indicators. Furthermore, in comparison to Transformer, \ours{} outperforms it on both \texttt{ROUGE-1} and \texttt{ROUGE-L}, while achieving comparable performance in terms of \texttt{ROUGE-2}. Notably, when the sample number is 500, \ours{} demonstrates superiority over Transformer across all the measures.

\begin{table}[ht]
\small 
\centering
\setlength\tabcolsep{10pt} 
\caption{Results on IWSLT14 \textsc{De}→\textsc{En} test set following the setting of \textsc{SeqDiffuSeq}. ``NFE'' indicates the \textbf{N}umber of \textbf{F}unction \textbf{E}valuations \citep{ye2023dinoiser}.}
\begin{tabular}{l|c|c|cc}
\toprule
\textbf{Methods} & \textbf{Pattern} &  \textbf{\texttt{BLEU}} & \textbf{Steps} & \textbf{NFE (Steps$\times k$)}  \\ \midrule
Transformer \citep{NIPS2017_3f5ee243} & AR & 34.74 & - & - \\   \midrule
CNAT \citep{bao-etal-2021-non} & NAR & 29.81 & - & - \\ \midrule
SeqDiffuSeq \citep{yuan2022seqdiffuseq} ($k$ = 1) & \multirow{2}{*}{Diffusion}  & 29.83 & 2,000 & 2,000 (2,000 $\times$ 1) \\
\ours{} ($k$ = 1) & & 30.19 & 20 & 20 (20 $\times$ 1) \\\midrule
GENIE \citep{lin2023text} ($k$ = 50) &  &  30.08 & 20 & 1,000 (20 $\times$ 50) \\ 
\ours{} ($k$ = 50) & Diffusion & \underline{34.95} & 20 & 1,000 (20 $\times$ 50) \\
\ours{} ($k$ = 500) & & \textbf{35.62} & 20 & 10,000 (20 $\times$ 500) \\
 \bottomrule
\end{tabular}
\label{seqdiffuseq main table}
\end{table}
\paragraph{Machine Translation}
\cref{seqdiffuseq main table} presents the \texttt{BLEU} score implemented by SeqDiffuSeq setting. \ours{} outperforms the non-auto-regressive CNAT in greedy search for a single sample, and achieves a substantial gain. Moreover, the \texttt{BLEU} score of \ours{} surpasses GENIE by a large margin and shows a slightly better performance than the AR Transformer. Specially, \ours{} achieves a more powerful result at $k$ = 500.

In \cref{translation main table} we give the \texttt{SacreBLEU} score according to the setting of DINOISER. \ours{} has notable improvements over non-auto-regressive CMLM. \,
Besides, \ours{} achieves excellent performance among text diffusion models for both \textsc{En}→\textsc{De} and \textsc{De}→\textsc{En} tasks. Specifically, \ours{} is far superior to GENIE and comparable to the newly proposed DINOISER at $n$ = 50. Nevertheless, the performance is stronger than DINOISER when $k$ = 500\footnote{DINOISER has shown in their Figure 4 that their method is not better with a larger $k$.}.

\begin{table}[ht]
\small 
\centering
\setlength\tabcolsep{14pt} 
\caption{\texttt{SacreBLEU} on the IWSLT14  test set. This result follows the setting of \textsc{DiNoiSer}.}
\begin{tabular}{l|cc}
\toprule
\multirow{2}{*}{\textbf{Methods}} & \multicolumn{2}{c}{IWSLT14}\\ \cmidrule{2-3} 
                         & \textsc{De}→\textsc{En}        & \textsc{En}→\textsc{De}  \\ \midrule
Transformer (AR, beam = 5) \citep{NIPS2017_3f5ee243} & 33.61 & 28.30 \\\midrule
CMLM (NAR, $k$ = 5) \citep{ghazvininejad-etal-2019-mask} & 29.41 & 24.34 \\\midrule
DiffusionLM ($k$ = 50) \citep{li2022diffusionlm} & 29.11 & 22.91  \\
DINOISER ($k$ = 50) \citep{ye2023dinoiser} & 31.61 & \underline{26.14}  \\\midrule
GENIE ($k$ = 50) \citep{lin2023text} & 29.45 & 23.89  \\
\ours{} ($k$ = 50) & \underline{31.80} & 26.01  \\
\ours{} ($k$ = 500) & \textbf{32.35} & \textbf{26.51} \\ \bottomrule
\end{tabular}
\label{translation main table}
\end{table}


\begin{table}[ht]
\small
\centering
\setlength\tabcolsep{5pt} 
\caption{Results on \textsc{CommonGen} dev set. Results of NAR and AR are from \cite{lin-etal-2020-commongen}.}
\begin{tabular}{l|c|cc|cc|cc}
\toprule
\textbf{Methods} &  \textbf{Pattern} & \multicolumn{2}{c}{\textbf{\texttt{ROUGE-2/L}}} & \multicolumn{2}{c}{\textbf{\texttt{BLEU-3/4}}} & \textbf{\texttt{METEOR}} & \textbf{\texttt{SPICE}} \\ \midrule
bRNN-CopyNet \citep{GuLLL16} & \multirow{3}{*}{AR} & 9.23 &  30.57 & 13.60 & 7.80 & 17.40 &  16.90 \\
Trans-CopyNet \citep{lin-etal-2020-commongen} & & 11.08 & 32.57 & 17.20 & 10.60 & 18.80 & 18.00 \\
MeanPooling-CopyNet \citep{lin-etal-2020-commongen} & & 11.36 & 34.63 & 14.80 & 8.90 & 19.20 & 20.20 \\\midrule
LevT \citep{gu2019levenshtein} & \multirow{2}{*}{NAR} & 12.22 & \underline{35.42} & \underline{23.10} & \underline{15.00} & 22.10 & 21.40 \\
ConstLeven \citep{SusantoCT20} & & \underline{13.47} & 35.19 & 21.30 & 12.30 &  \textbf{25.00} & \underline{ 23.20} \\\midrule
GENIE \citep{lin2023text} ($k$ = 50) & \multirow{2}{*}{Diffusion} & 12.89 & 35.21 & 22.00 & 13.30 & \underline{24.30} & 23.00 \\
\ours{} ($k$ = 50) & & \textbf{13.93} & \textbf{37.36} & \textbf{25.60} & \textbf{16.40} & \textbf{25.00} & \textbf{24.20} \\
\bottomrule
\end{tabular}
\label{commongen main table}
\end{table}
\paragraph{Common Sense Generation} 
As depicted in \cref{commongen main table}, \ours{} achieves superior performance compared to the current AR, NAR, and other diffusion methods across all the metrics on the \textsc{CommonGen} dataset. 

\begin{table}[ht]
\small 
\centering
\setlength\tabcolsep{8pt} 
\caption{Experimental results of GENIE and \ours{} with inference steps of \textbf{2} and \textbf{3} on \textsc{XSum} test set. Take $k$ = 10 to apply the MBR decoding strategy. $(\cdot)$ indicates the \textbf{drop} score compared to the 20-step.}
\begin{tabular}{l|cc|ccc|c}
\toprule
\textbf{Methods}                   & \textbf{Steps} & \textbf{NFE} & \textbf{\texttt{ROUGE-1}} & \textbf{\texttt{ROUGE-2}} & \textbf{\texttt{ROUGE-L}} & \textbf{AVG Drop} \\\midrule
\multirow{5}{*}{GENIE}    & 2,000 & 20,000 & 30.36  & 8.78   & 23.31  & - \\\cmidrule{2-7}
                          & 20 & 200 & 28.33  & 7.46   & 21.15  & - \\\cmidrule{2-7}
                          & 3 & 30 & 25.03 (-3.30) & 5.32 (-2.14) & 18.17 (-2.98) & 2.81     \\
                          & 2 & 20 & 23.45 (-4.88) & 3.95 (-3.51) & 16.94 (-4.21) & 4.20    \\\midrule
\multirow{3}{*}{\ours{}} & 20 & 200 & 30.99  & 9.32   & 23.95  & -  \\\cmidrule{2-7}
                          & 3 & 30 & 30.23 (-0.76) & 8.68 (-0.64) & 23.43 (-0.52) & \textbf{0.64}     \\ 
                          & 2 & 20 & 29.28 (-1.71) & 7.99 (-1.33) & 22.98 (-0.97) & \textbf{1.34}    \\\bottomrule
\end{tabular}
\label{table:fewer_steps}
\end{table}
\subsection{Inference Efficiency}

First, we use the number of function evaluations (NFE) as a measure to compare inference efficiency~\citep{ye2023dinoiser} in machine translation. From Table \ref{seqdiffuseq main table}, it is evident that even when the NFE is reduced to 1\% of SeqDiffuSeq (equivalent to $100\times$ faster), \ours{} still outperforms SeqDiffuSeq. Moreover, increasing the number of generated candidate samples ($k=500$) leads to further performance improvements, albeit with increased time consumption.

Second, we conduct experiments with an \textbf{extremely limited number of inference steps} (2 and 3)\footnote{The time consumed by each step in the inference process is exactly the same.} and compare the performance with that of GENIE in \textsc{XSum}. The results are presented in \cref{table:fewer_steps}. When reducing the number of steps to 2, GENIE experiences a significant decline, with an average score of 4.20 in the AVG Drop column, while \ours{} exhibits a comparatively smaller decrease of 1.34. Furthermore, with 3 steps, although the performance deterioration of GENIE is somewhat reduced, the average score still shows a decline of 2.81. In contrast, \ours{} maintains a high performance level, with an average score differing from the 20-step result by only 0.64. Notably, the results of \ours{} at 3 steps are comparable to the results of GENIE at 2,000 steps. Therefore, compared to GENIE, the inference speed of \ours{} can be accelerated by up to $600\times$.

\begin{table}[ht]
\small
\centering
\captionof{table}{Diversity of \textbf{10} generated samples on \textsc{XSum} test set and average of \textbf{10} results. The results of BART and GENIE are quoted from \cite{lin2023text}.}
\setlength\tabcolsep{4.5pt} 
\begin{tabular}{c|cccccc|cc}
\toprule
\textbf{Methods} & \multicolumn{6}{c|}{BART} & GENIE & \ours{} \\ \midrule
\textbf{Sampling} & \begin{tabular}[c]{@{}c@{}}Greedy\\ Search\end{tabular} & \begin{tabular}[c]{@{}c@{}}Beam\\ Search\end{tabular} & \begin{tabular}[c]{@{}c@{}}Diverse\\ Beam Search\end{tabular} & \begin{tabular}[c]{@{}c@{}}Typical\\ Sample\end{tabular} & \begin{tabular}[c]{@{}c@{}}Top-k\\ Sample\end{tabular} & \begin{tabular}[c]{@{}c@{}}Nucleus\\ Sample\end{tabular} & \multicolumn{2}{c}{Diffusion} \\\midrule
\textbf{\texttt{SELF-BLEU} ↓} & 100.0 & 93.4 & 75.6 & 76.9 & 80.2 & 79.1 & 29.3 & 30.4 \\
 \bottomrule
\end{tabular}
\label{diversity}
\end{table}

\subsection{Analysis}

\paragraph{Diversity of Samples}
Diversity is a key advantage of diffusion models. To measure the diversity of generated samples, We adopt the \texttt{SELF-BLEU} \citep{zhu2018texygen} metric, in which a lower score indicates higher diversity. In \cite{lin2023text}, various sampling methods were applied to the pre-trained auto-regressive model BART, including Greedy Search, Beam Search~\cite{journals/corr/abs-2204-09269}, Diverse Beam Search(diversity strength = 0.8)~\cite{journals/corr/VijayakumarCSSL16}, Typical Sample ($\tau$ = 1.2)~\cite{journals/corr/abs-2202-00666}, Top-k Sample ($k$ = 50)~\cite{LewisDF18} and Nucleus Sample ($p$ = 0.92)~\cite{HoltzmanBDFC20}. 

Specifically, greedy search is to select the token with the highest probability at each step. Beam search is to select the largest token from among the beams with higher probability at each step. Diverse beam search is to divide the beams into multiple groups at each step and ensure the difference between groups by calculating the diversity score between groups.  Typical sampling selects samples through a discrete random process. Top-k sampling is to randomly select one of the $k$ candidate tokens with the highest probability at each step. Nucleus sampling is to randomly select one token at each step from the candidate tokens whose probability density is greater than $p$. 

As shown in \cref{diversity}, \ours{} achieves significantly higher diversity compared to the auto-regressive model. Furthermore, the diversity can be comparable to GENIE with a better performance.

\begin{figure}[ht]
\centering
\subfigure[]{
\begin{minipage}[t]{0.316\linewidth}
\label{ablation_train}
\centering
\includegraphics[width=1.7in]{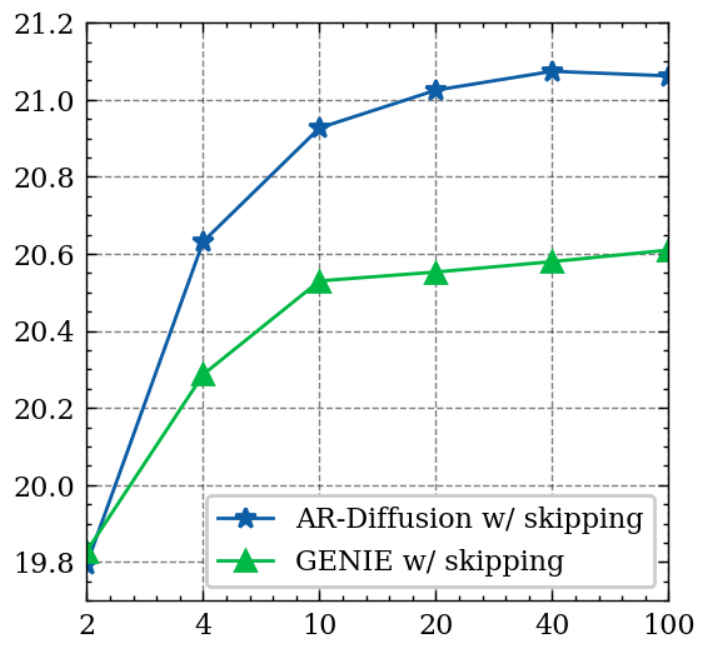}
\end{minipage}
}
\subfigure[]{
\begin{minipage}[t]{0.316\linewidth}
\label{ablation_infer}
\centering
\includegraphics[width=1.7in]{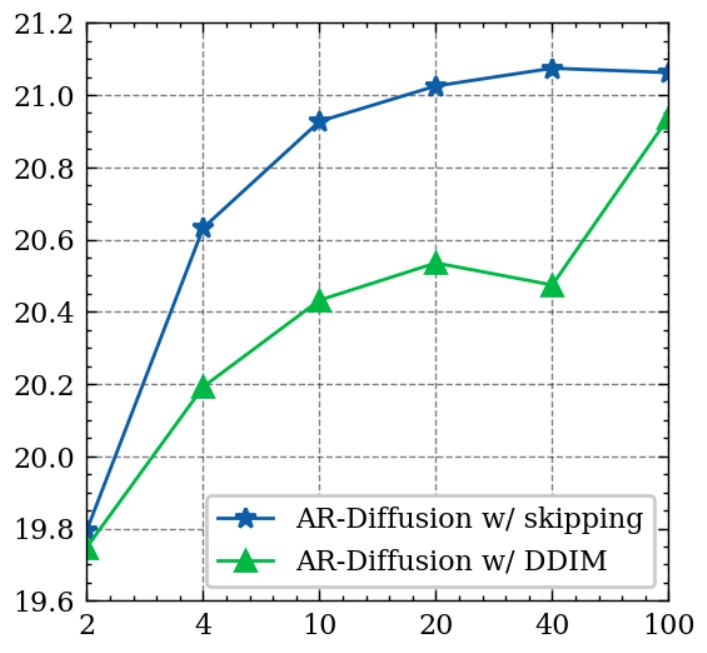}
\end{minipage}
}
\subfigure[]{
\begin{minipage}[t]{0.316\linewidth}
\label{ablation_baseline}
\centering
\includegraphics[width=1.7in]{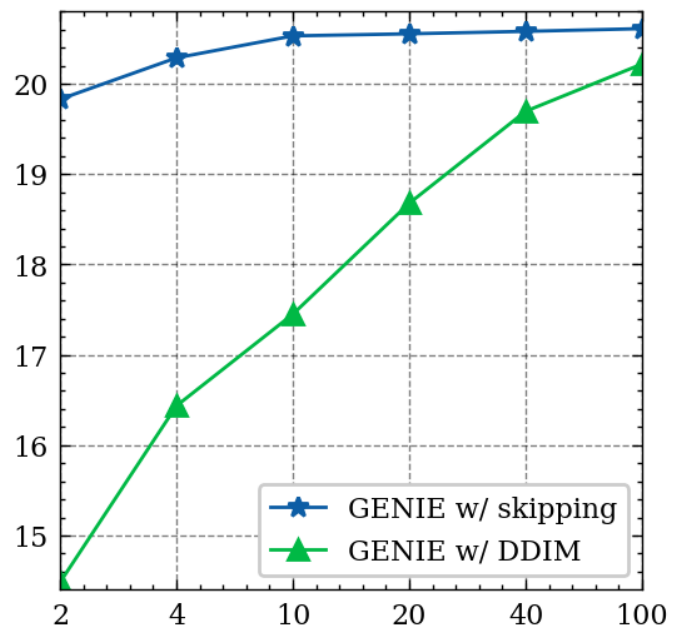}
\end{minipage}
}
\centering
\caption{Ablation experiments on \textsc{XSum} test set and taking $k$ = 5. The horizontal axis is the number of inference steps and the vertical axis is \texttt{AVG-ROUGE} = (\texttt{ROUGE-1} + \texttt{ROUGE-2} + \texttt{ROUGE-L}) / 3.}
\end{figure}

\paragraph{Ablation Study}
To demonstrate the effectiveness of our proposed method, we perform ablation experiments on the \textsc{XSum} dataset. Our results show that both our proposed multi-level diffusion and skipping mechanism are essential for achieving the high performance of \ours{}.

Maintaining the skipping inference method, we remove the token-level diffusion during the training process, which degenerates to GENIE w/ skipping. The comparison results are shown in \cref{ablation_train}.  It can be observed that after removing, the \texttt{AVG-ROUGE} score is greatly lower after 2 steps. \looseness=-1

The performance of applying our proposed skipping mechanism and DDIM \citep{DBLP:conf/iclr/SongME21} to \ours{} is shown in \cref{ablation_infer}. The results demonstrate that the skipping mechanism consistently outperforms DDIM in various inference steps. Additionally, the skipping mechanism can be easily applied to GENIE. As depicted in \cref{ablation_baseline}, DDIM suffers a significant drop in performance when the number of inference steps is less than 40. In contrast, the skipping mechanism consistently maintains good performance across all inference steps.

\begin{figure}[ht]
\centering
\includegraphics[width=\linewidth]{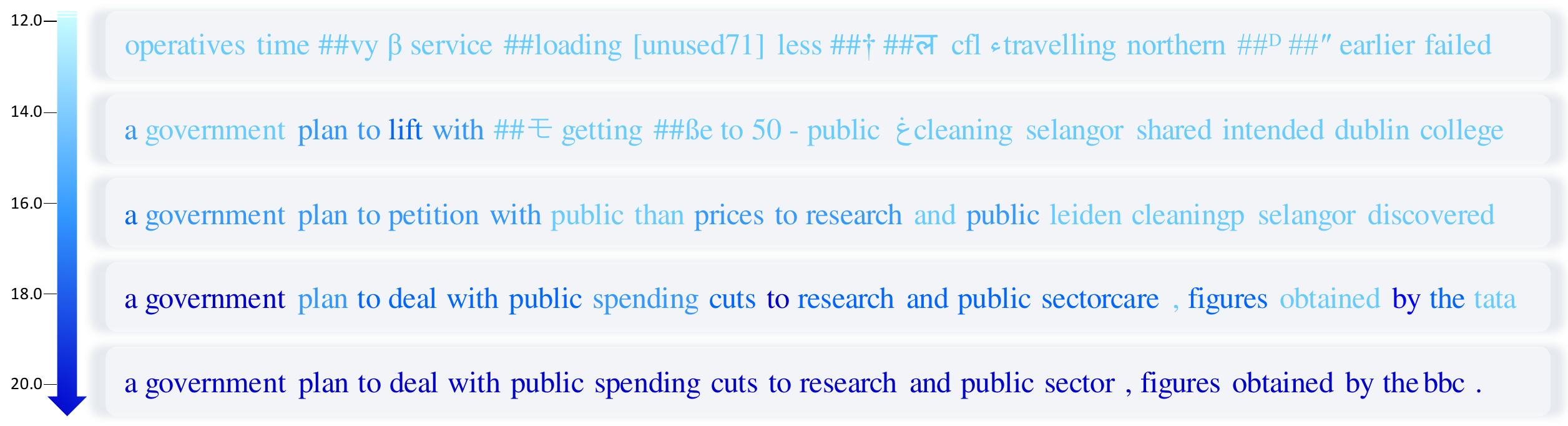}
\caption{The intermediate state of \ours{} gradually generating real text from a standard Gaussian noise through 20 steps. The brightness of the color represents the magnitude of the logits, with darker colors indicating larger logits. More cases are shown in the supplementary materials \ref{appendix:cases}.}
\label{case}
\end{figure}

\paragraph{Case Study}
By mapping the state to the token with the highest logits, we visualize the intermediate states of \ours{}. As depicted in \cref{case}, \ours{} undergoes a denoising process, transforming the random Gaussian noise into a coherent sentence over 20 steps, and we present 5 of them. With the progression of each timestep, compared to the tokens on the right side of the sentence, the tokens on the left side demonstrate faster determination and a rapid increase in the corresponding logits. This behavior is consistent with our principle of dynamic movement speed from left to right. \looseness=-1

\subsection{Impact of Minimum Bayes Risk and Anchor Point}

\begin{figure}[ht]
\begin{minipage}[bp]{0.5\textwidth} 
\centering 
\includegraphics[width=1.0\textwidth]{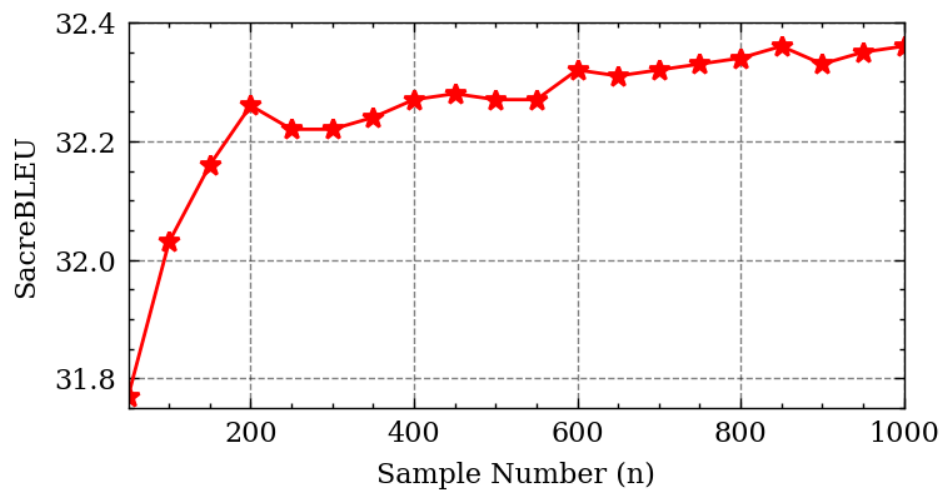}
\captionsetup{width=0.9\textwidth}
\caption{Relationship between the number of candidate samples for applying MBR and \texttt{SacreBLEU} on IWSLT14 \textsc{De}→\textsc{En} test set.} 
\label{mbr}
\end{minipage}
\begin{minipage}[bp]{0.5\textwidth} 
\small
\centering
\captionof{table}{Effect of anchor point at different positions on the IWSLT14 \textsc{De}→\textsc{En} test set. ``$N$'' indicates the target sequence length and ``T'' represents the total time step of diffusion.}
\setlength\tabcolsep{3pt} 
\begin{tabular}{cc|c}
\toprule
$n_e$ & $t_e$ & \texttt{SacreBLEU}   \\ \midrule
$1.0 \times N$ & $T$ & 31.23  \\
$2.0 \times N$ & $T$ & \textbf{31.80} \\ 
$3.0 \times N$ & $T$ & 31.58\\
\bottomrule
\end{tabular}
\label{anchor_N}
\end{minipage} 
\end{figure}

\paragraph{Minimum Bayes Risk}
To investigate the relationship between the number of generated candidate samples ($k$) and the quality of generation, we generate varying numbers of samples, ranging up to 1,000, on the IWSLT14 De→En test set and present the results in \cref{mbr}. The curve demonstrates an initial gain of approximately 0.5 \texttt{SacreBLEU} within the first 200 samples, after which the gain becomes insignificant with generating more samples.

\paragraph{Anchor Point}
We conduct experiments on $\ours$ using different anchor points $(n_e, t_e)$. These anchor points vary in terms of $n_e$ values, namely {$1.0 \times N$, $2.0 \times N$ and $3.0 \times N$}, where $N$ denotes the target sentence length. Additionally, they share a common $t_e$ value of $T$, which represents the total time step of diffusion. We present the results in \cref{anchor_N}, and determine that the best result is achieved at $(n_e, t_e)$ = $(2.0 \times N, T)$.

\section{Related Work}
\paragraph{AR and NAR Language Models}

AR models have been the dominant approach for text generation \citep{DBLP:journals/corr/abs-2303-08774,DBLP:journals/corr/abs-2302-13971,DBLP:journals/csur/DongLGCLSY23}, but their token-by-token generation nature often leads to unsatisfactory inference speed. To address this issue, NAR models have been developed in recent years. The NAR method is initially proposed by \cite{gu2017non}, its objective is generate the entire output sequence in parallel, thereby improving generation speed and efficiency. Subsequently, LevT \citep{gu2019levenshtein} adopts insertion and deletion to address the lack of flexibility in NAR generation, CMLM \citep{ghazvininejad-etal-2019-mask} utilizes a masked language model to improve the quality of NAR generation through a constant number of iterations, and CNAT \citep{bao-etal-2021-non} introduces latent variables to represent the category information of the target word to make full use of the latent representation. However, these NAR methods are hard to model inter-token position dependency and deficient in generation performance.

\paragraph{Continuous Text Diffusion}

The application of diffusion models to continuous text space is first introduced by \cite{li2022diffusionlm}.  Through the embedding and rounding processes, the direct integration of continuous noise into word embeddings was accomplished. After that, more people attempt to adopt continuous text diffusion model to solve sequence-to-sequence tasks. DiffuSeq \citep{DBLP:journals/corr/abs-2210-08933} divides the input into two parts, utilizing one part as a condition, and perturbs the other part with noise. CDCD \citep{dieleman2022continuous} proposes score interpolation and time warping to allow diffusion model and Euclidean embedding to share the same loss function for training. SeqDiffuSeq \citep{yuan2022seqdiffuseq}, GENIE \citep{lin2023text} and DINOISER \citep{ye2023dinoiser} incorporate diffusion model into the encoder-decoder structure through cross-attention mechanisms. 

It is important to highlight the differences between our method and both ARDMs \citep{hoogeboom2022autoregressive} and TimeGrad \citep{pmlr-v139-rasul21a}, despite the common references to autoregression and diffusion in all these. ARDMs employ an order-agnostic technique, leveraging masking and prediction for generation in arbitrary orders. On the other hand, TimeGrad integrates RNN and diffusion to model the conditional distribution of future steps of multivariate time series. In contrast, our research focuses on implementing the diffusion process within a continuous embedding space, with the primary aim of generating text in a left-to-right sequence.

\section{Conclusion}
This paper introduces \ours{}, which exhibits AR-like generation behavior but enables efficient parallel decoding. Embracing the inherent sequential nature of language, we propose a multi-level diffusion model, consisting of sentence-level and token-level components, to assign dynamic movement speeds to tokens. Consequently, compared to those on the right, the left tokens undergo fewer denoising steps and generate earlier to subsequently influence the later ones. Furthermore, we introduce a skipping mechanism to facilitate parallel generation within the multi-level diffusion framework. The experimental results across various tasks demonstrate that \ours{} surpasses existing diffusion models in terms of quality while maintaining diversity. Additionally, compared to existing diffusion language models, \ours{} achieves comparable results while being $100\times\sim 600\times$ faster.

\section{Limitation}
A primary limitation of our work lies in the requirement of generating a large number of candidate samples for optimal performance. As an illustration in \cref{cnndm main table} of \textsc{Cnn/DailyMail} dataset, \ours{} ($k$ = 50) achieves a 0.8 lower \texttt{ROUGE-2} score compared to \ours{} ($k$ = 500). We anticipate exploring more efficient sampling strategies to minimize the number of generated samples without performance drop.

\bibliographystyle{unsrtnat}
\bibliography{ref}

\begin{thebibliography}{43}
\providecommand{\natexlab}[1]{#1}
\providecommand{\url}[1]{\texttt{#1}}
\expandafter\ifx\csname urlstyle\endcsname\relax
  \providecommand{\doi}[1]{doi: #1}\else
  \providecommand{\doi}{doi: \begingroup \urlstyle{rm}\Url}\fi

\bibitem[OpenAI(2023)]{DBLP:journals/corr/abs-2303-08774}
OpenAI.
\newblock {GPT-4} technical report.
\newblock \emph{CoRR}, abs/2303.08774, 2023.
\newblock \doi{10.48550/arXiv.2303.08774}.
\newblock URL \url{https://doi.org/10.48550/arXiv.2303.08774}.

\bibitem[Touvron et~al.(2023)Touvron, Lavril, Izacard, Martinet, Lachaux,
  Lacroix, Rozi{\`{e}}re, Goyal, Hambro, Azhar, Rodriguez, Joulin, Grave, and
  Lample]{DBLP:journals/corr/abs-2302-13971}
Hugo Touvron, Thibaut Lavril, Gautier Izacard, Xavier Martinet, Marie{-}Anne
  Lachaux, Timoth{\'{e}}e Lacroix, Baptiste Rozi{\`{e}}re, Naman Goyal, Eric
  Hambro, Faisal Azhar, Aur{\'{e}}lien Rodriguez, Armand Joulin, Edouard Grave,
  and Guillaume Lample.
\newblock Llama: Open and efficient foundation language models.
\newblock \emph{CoRR}, abs/2302.13971, 2023.
\newblock \doi{10.48550/arXiv.2302.13971}.
\newblock URL \url{https://doi.org/10.48550/arXiv.2302.13971}.

\bibitem[Taori et~al.(2023)Taori, Gulrajani, Zhang, Dubois, Li, Guestrin,
  Liang, and Hashimoto]{alpaca}
Rohan Taori, Ishaan Gulrajani, Tianyi Zhang, Yann Dubois, Xuechen Li, Carlos
  Guestrin, Percy Liang, and Tatsunori~B. Hashimoto.
\newblock Stanford alpaca: An instruction-following llama model.
\newblock \url{https://github.com/tatsu-lab/stanford_alpaca}, 2023.

\bibitem[Vaswani et~al.(2017)Vaswani, Shazeer, Parmar, Uszkoreit, Jones, Gomez,
  Kaiser, and Polosukhin]{NIPS2017_3f5ee243}
Ashish Vaswani, Noam Shazeer, Niki Parmar, Jakob Uszkoreit, Llion Jones,
  Aidan~N Gomez, \L~ukasz Kaiser, and Illia Polosukhin.
\newblock Attention is all you need.
\newblock In I.~Guyon, U.~Von Luxburg, S.~Bengio, H.~Wallach, R.~Fergus,
  S.~Vishwanathan, and R.~Garnett, editors, \emph{Advances in Neural
  Information Processing Systems}, volume~30. Curran Associates, Inc., 2017.
\newblock URL
  \url{https://proceedings.neurips.cc/paper_files/paper/2017/file/3f5ee243547dee91fbd053c1c4a845aa-Paper.pdf}.

\bibitem[Brown et~al.(2020)Brown, Mann, Ryder, Subbiah, Kaplan, Dhariwal,
  Neelakantan, Shyam, Sastry, Askell, et~al.]{brown2020language}
Tom Brown, Benjamin Mann, Nick Ryder, Melanie Subbiah, Jared~D Kaplan, Prafulla
  Dhariwal, Arvind Neelakantan, Pranav Shyam, Girish Sastry, Amanda Askell,
  et~al.
\newblock Language models are few-shot learners.
\newblock \emph{Advances in neural information processing systems},
  33:\penalty0 1877--1901, 2020.

\bibitem[Ho et~al.(2020)Ho, Jain, and Abbeel]{DBLP:conf/nips/HoJA20}
Jonathan Ho, Ajay Jain, and Pieter Abbeel.
\newblock Denoising diffusion probabilistic models.
\newblock In Hugo Larochelle, Marc'Aurelio Ranzato, Raia Hadsell,
  Maria{-}Florina Balcan, and Hsuan{-}Tien Lin, editors, \emph{Advances in
  Neural Information Processing Systems 33: Annual Conference on Neural
  Information Processing Systems 2020, NeurIPS 2020, December 6-12, 2020,
  virtual}, 2020.
\newblock URL
  \url{https://proceedings.neurips.cc/paper/2020/hash/4c5bcfec8584af0d967f1ab10179ca4b-Abstract.html}.

\bibitem[Li et~al.(2022{\natexlab{a}})Li, Thickstun, Gulrajani, Liang, and
  Hashimoto]{li2022diffusionlm}
Xiang~Lisa Li, John Thickstun, Ishaan Gulrajani, Percy Liang, and Tatsunori
  Hashimoto.
\newblock Diffusion-{LM} improves controllable text generation.
\newblock In Alice~H. Oh, Alekh Agarwal, Danielle Belgrave, and Kyunghyun Cho,
  editors, \emph{Advances in Neural Information Processing Systems},
  2022{\natexlab{a}}.
\newblock URL \url{https://openreview.net/forum?id=3s9IrEsjLyk}.

\bibitem[Gong et~al.(2022)Gong, Li, Feng, Wu, and
  Kong]{DBLP:journals/corr/abs-2210-08933}
Shansan Gong, Mukai Li, Jiangtao Feng, Zhiyong Wu, and Lingpeng Kong.
\newblock Diffuseq: Sequence to sequence text generation with diffusion models.
\newblock \emph{CoRR}, abs/2210.08933, 2022.
\newblock \doi{10.48550/arXiv.2210.08933}.
\newblock URL \url{https://doi.org/10.48550/arXiv.2210.08933}.

\bibitem[Dieleman et~al.(2022)Dieleman, Sartran, Roshannai, Savinov, Ganin,
  Richemond, Doucet, Strudel, Dyer, Durkan, et~al.]{dieleman2022continuous}
Sander Dieleman, Laurent Sartran, Arman Roshannai, Nikolay Savinov, Yaroslav
  Ganin, Pierre~H Richemond, Arnaud Doucet, Robin Strudel, Chris Dyer, Conor
  Durkan, et~al.
\newblock Continuous diffusion for categorical data.
\newblock \emph{arXiv preprint arXiv:2211.15089}, 2022.

\bibitem[Yuan et~al.(2022)Yuan, Yuan, Tan, Huang, and
  Huang]{yuan2022seqdiffuseq}
Hongyi Yuan, Zheng Yuan, Chuanqi Tan, Fei Huang, and Songfang Huang.
\newblock Seqdiffuseq: Text diffusion with encoder-decoder transformers, 2022.

\bibitem[Ye et~al.(2023)Ye, Zheng, Bao, Qian, and Wang]{ye2023dinoiser}
Jiasheng Ye, Zaixiang Zheng, Yu~Bao, Lihua Qian, and Mingxuan Wang.
\newblock Dinoiser: Diffused conditional sequence learning by manipulating
  noises, 2023.

\bibitem[Lewis et~al.(2020)Lewis, Liu, Goyal, Ghazvininejad, Mohamed, Levy,
  Stoyanov, and Zettlemoyer]{DBLP:conf/acl/LewisLGGMLSZ20}
Mike Lewis, Yinhan Liu, Naman Goyal, Marjan Ghazvininejad, Abdelrahman Mohamed,
  Omer Levy, Veselin Stoyanov, and Luke Zettlemoyer.
\newblock {BART:} denoising sequence-to-sequence pre-training for natural
  language generation, translation, and comprehension.
\newblock In Dan Jurafsky, Joyce Chai, Natalie Schluter, and Joel~R. Tetreault,
  editors, \emph{Proceedings of the 58th Annual Meeting of the Association for
  Computational Linguistics, {ACL} 2020, Online, July 5-10, 2020}, pages
  7871--7880. Association for Computational Linguistics, 2020.
\newblock \doi{10.18653/v1/2020.acl-main.703}.
\newblock URL \url{https://doi.org/10.18653/v1/2020.acl-main.703}.

\bibitem[Qi et~al.(2020)Qi, Yan, Gong, Liu, Duan, Chen, Zhang, and
  Zhou]{qi-etal-2020-prophetnet}
Weizhen Qi, Yu~Yan, Yeyun Gong, Dayiheng Liu, Nan Duan, Jiusheng Chen, Ruofei
  Zhang, and Ming Zhou.
\newblock {P}rophet{N}et: Predicting future n-gram for
  sequence-to-{S}equence{P}re-training.
\newblock In \emph{Findings of the Association for Computational Linguistics:
  EMNLP 2020}, pages 2401--2410, Online, November 2020. Association for
  Computational Linguistics.
\newblock \doi{10.18653/v1/2020.findings-emnlp.217}.
\newblock URL \url{https://aclanthology.org/2020.findings-emnlp.217}.

\bibitem[Qi et~al.(2021)Qi, Gong, Jiao, Yan, Chen, Liu, Tang, Li, Chen, Zhang,
  et~al.]{qi2021bang}
Weizhen Qi, Yeyun Gong, Jian Jiao, Yu~Yan, Weizhu Chen, Dayiheng Liu, Kewen
  Tang, Houqiang Li, Jiusheng Chen, Ruofei Zhang, et~al.
\newblock Bang: Bridging autoregressive and non-autoregressive generation with
  large scale pretraining.
\newblock In \emph{International Conference on Machine Learning}, pages
  8630--8639. PMLR, 2021.

\bibitem[Li et~al.(2022{\natexlab{b}})Li, Tang, Zhao, Nie, and
  Wen]{DBLP:conf/emnlp/LiTZNW22}
Junyi Li, Tianyi Tang, Wayne~Xin Zhao, Jian{-}Yun Nie, and Ji{-}Rong Wen.
\newblock {ELMER:} {A} non-autoregressive pre-trained language model for
  efficient and effective text generation.
\newblock In Yoav Goldberg, Zornitsa Kozareva, and Yue Zhang, editors,
  \emph{Proceedings of the 2022 Conference on Empirical Methods in Natural
  Language Processing, {EMNLP} 2022, Abu Dhabi, United Arab Emirates, December
  7-11, 2022}, pages 1044--1058. Association for Computational Linguistics,
  2022{\natexlab{b}}.
\newblock URL \url{https://aclanthology.org/2022.emnlp-main.68}.

\bibitem[Li et~al.(2022{\natexlab{c}})Li, Cui, Yin, and
  Zhang]{DBLP:conf/emnlp/LiCY022}
Yafu Li, Leyang Cui, Yongjing Yin, and Yue Zhang.
\newblock Multi-granularity optimization for non-autoregressive translation.
\newblock In Yoav Goldberg, Zornitsa Kozareva, and Yue Zhang, editors,
  \emph{Proceedings of the 2022 Conference on Empirical Methods in Natural
  Language Processing, {EMNLP} 2022, Abu Dhabi, United Arab Emirates, December
  7-11, 2022}, pages 5073--5084. Association for Computational Linguistics,
  2022{\natexlab{c}}.
\newblock URL \url{https://aclanthology.org/2022.emnlp-main.339}.

\bibitem[Bao et~al.(2021)Bao, Huang, Xiao, Wang, Dai, and
  Chen]{bao-etal-2021-non}
Yu~Bao, Shujian Huang, Tong Xiao, Dongqi Wang, Xinyu Dai, and Jiajun Chen.
\newblock Non-autoregressive translation by learning target categorical codes.
\newblock In \emph{Proceedings of the 2021 Conference of the North American
  Chapter of the Association for Computational Linguistics: Human Language
  Technologies}, pages 5749--5759, Online, June 2021. Association for
  Computational Linguistics.
\newblock \doi{10.18653/v1/2021.naacl-main.458}.
\newblock URL \url{https://aclanthology.org/2021.naacl-main.458}.

\bibitem[Lin et~al.(2023)Lin, Gong, Shen, Wu, Fan, Lin, Duan, and
  Chen]{lin2023text}
Zhenghao Lin, Yeyun Gong, Yelong Shen, Tong Wu, Zhihao Fan, Chen Lin, Nan Duan,
  and Weizhu Chen.
\newblock Text generation with diffusion language models: A pre-training
  approach with continuous paragraph denoise, 2023.

\bibitem[Gu et~al.(2017)Gu, Bradbury, Xiong, Li, and Socher]{gu2017non}
Jiatao Gu, James Bradbury, Caiming Xiong, Victor~OK Li, and Richard Socher.
\newblock Non-autoregressive neural machine translation.
\newblock \emph{arXiv preprint arXiv:1711.02281}, 2017.

\bibitem[Narayan et~al.(2018)Narayan, Cohen, and
  Lapata]{narayan-etal-2018-dont}
Shashi Narayan, Shay~B. Cohen, and Mirella Lapata.
\newblock Don{'}t give me the details, just the summary! topic-aware
  convolutional neural networks for extreme summarization.
\newblock In \emph{Proceedings of the 2018 Conference on Empirical Methods in
  Natural Language Processing}, pages 1797--1807, Brussels, Belgium,
  October-November 2018. Association for Computational Linguistics.
\newblock \doi{10.18653/v1/D18-1206}.
\newblock URL \url{https://aclanthology.org/D18-1206}.

\bibitem[Hermann et~al.(2015)Hermann, Kocisky, Grefenstette, Espeholt, Kay,
  Suleyman, and Blunsom]{NIPS2015_afdec700}
Karl~Moritz Hermann, Tomas Kocisky, Edward Grefenstette, Lasse Espeholt, Will
  Kay, Mustafa Suleyman, and Phil Blunsom.
\newblock Teaching machines to read and comprehend.
\newblock In C.~Cortes, N.~Lawrence, D.~Lee, M.~Sugiyama, and R.~Garnett,
  editors, \emph{Advances in Neural Information Processing Systems}, volume~28.
  Curran Associates, Inc., 2015.
\newblock URL
  \url{https://proceedings.neurips.cc/paper_files/paper/2015/file/afdec7005cc9f14302cd0474fd0f3c96-Paper.pdf}.

\bibitem[Kudo and Richardson(2018)]{kudo-richardson-2018-sentencepiece}
Taku Kudo and John Richardson.
\newblock {S}entence{P}iece: A simple and language independent subword
  tokenizer and detokenizer for neural text processing.
\newblock In \emph{Proceedings of the 2018 Conference on Empirical Methods in
  Natural Language Processing: System Demonstrations}, pages 66--71, Brussels,
  Belgium, November 2018. Association for Computational Linguistics.
\newblock \doi{10.18653/v1/D18-2012}.
\newblock URL \url{https://aclanthology.org/D18-2012}.

\bibitem[Lee et~al.(2018)Lee, Mansimov, and Cho]{lee2018deterministic}
Jason Lee, Elman Mansimov, and Kyunghyun Cho.
\newblock Deterministic non-autoregressive neural sequence modeling by
  iterative refinement.
\newblock \emph{arXiv preprint arXiv:1802.06901}, 2018.

\bibitem[Ghazvininejad et~al.(2019)Ghazvininejad, Levy, Liu, and
  Zettlemoyer]{ghazvininejad-etal-2019-mask}
Marjan Ghazvininejad, Omer Levy, Yinhan Liu, and Luke Zettlemoyer.
\newblock Mask-predict: Parallel decoding of conditional masked language
  models.
\newblock In \emph{Proceedings of the 2019 Conference on Empirical Methods in
  Natural Language Processing and the 9th International Joint Conference on
  Natural Language Processing (EMNLP-IJCNLP)}, pages 6112--6121, Hong Kong,
  China, November 2019. Association for Computational Linguistics.
\newblock \doi{10.18653/v1/D19-1633}.
\newblock URL \url{https://aclanthology.org/D19-1633}.

\bibitem[Gu et~al.(2019)Gu, Wang, and Zhao]{gu2019levenshtein}
Jiatao Gu, Changhan Wang, and Junbo Zhao.
\newblock Levenshtein transformer.
\newblock In \emph{Advances in Neural Information Processing Systems}, pages
  11181--11191, 2019.

\bibitem[Stern et~al.(2019)Stern, Chan, Kiros, and
  Uszkoreit]{stern2019insertion}
Mitchell Stern, William Chan, Jamie Kiros, and Jakob Uszkoreit.
\newblock Insertion transformer: Flexible sequence generation via insertion
  operations.
\newblock \emph{arXiv preprint arXiv:1902.03249}, 2019.

\bibitem[Gu et~al.(2016)Gu, Lu, Li, and Li]{GuLLL16}
Jiatao Gu, Zhengdong Lu, Hang Li, and Victor O.~K. Li.
\newblock Incorporating copying mechanism in sequence-to-sequence learning.
\newblock In \emph{{ACL} {(1)}}. The Association for Computer Linguistics,
  2016.

\bibitem[Greff et~al.(2017)Greff, Srivastava, Koutn{\'{\i}}k, Steunebrink, and
  Schmidhuber]{GreffSKSS17}
Klaus Greff, Rupesh~Kumar Srivastava, Jan Koutn{\'{\i}}k, Bas~R. Steunebrink,
  and J{\"{u}}rgen Schmidhuber.
\newblock {LSTM:} {A} search space odyssey.
\newblock \emph{{IEEE} Trans. Neural Networks Learn. Syst.}, 28\penalty0
  (10):\penalty0 2222--2232, 2017.

\bibitem[Lin et~al.(2020)Lin, Zhou, Shen, Zhou, Bhagavatula, Choi, and
  Ren]{lin-etal-2020-commongen}
Bill~Yuchen Lin, Wangchunshu Zhou, Ming Shen, Pei Zhou, Chandra Bhagavatula,
  Yejin Choi, and Xiang Ren.
\newblock {C}ommon{G}en: A constrained text generation challenge for generative
  commonsense reasoning.
\newblock In \emph{Findings of the Association for Computational Linguistics:
  EMNLP 2020}, pages 1823--1840, Online, November 2020. Association for
  Computational Linguistics.
\newblock \doi{10.18653/v1/2020.findings-emnlp.165}.
\newblock URL \url{https://aclanthology.org/2020.findings-emnlp.165}.

\bibitem[Kumar and Byrne(2004)]{kumar2004minimum}
Shankar Kumar and William Byrne.
\newblock Minimum bayes-risk decoding for statistical machine translation.
\newblock Technical report, JOHNS HOPKINS UNIV BALTIMORE MD CENTER FOR LANGUAGE
  AND SPEECH PROCESSING (CLSP), 2004.

\bibitem[Liu et~al.(2021)Liu, Yan, Gong, Qi, Zhang, Jiao, Chen, Fu, Shou, Gong,
  et~al.]{liu2021glge}
Dayiheng Liu, Yu~Yan, Yeyun Gong, Weizhen Qi, Hang Zhang, Jian Jiao, Weizhu
  Chen, Jie Fu, Linjun Shou, Ming Gong, et~al.
\newblock Glge: A new general language generation evaluation benchmark.
\newblock In \emph{Findings of the Association for Computational Linguistics:
  ACL-IJCNLP 2021}, pages 408--420, 2021.

\bibitem[Susanto et~al.(2020)Susanto, Chollampatt, and Tan]{SusantoCT20}
Raymond~Hendy Susanto, Shamil Chollampatt, and Liling Tan.
\newblock Lexically constrained neural machine translation with levenshtein
  transformer.
\newblock In \emph{{ACL}}, pages 3536--3543. Association for Computational
  Linguistics, 2020.

\bibitem[Zhu et~al.(2018)Zhu, Lu, Zheng, Guo, Zhang, Wang, and
  Yu]{zhu2018texygen}
Yaoming Zhu, Sidi Lu, Lei Zheng, Jiaxian Guo, Weinan Zhang, Jun Wang, and Yong
  Yu.
\newblock Texygen: A benchmarking platform for text generation models.
\newblock In \emph{The 41st international ACM SIGIR conference on research \&
  development in information retrieval}, pages 1097--1100, 2018.

\bibitem[Xiao et~al.(2022)Xiao, Wu, Guo, Li, Zhang, Qin, and
  Liu]{journals/corr/abs-2204-09269}
Yisheng Xiao, Lijun Wu, Junliang Guo, Juntao Li, Min Zhang, Tao Qin, and
  Tie{-}Yan Liu.
\newblock A survey on non-autoregressive generation for neural machine
  translation and beyond.
\newblock \emph{CoRR}, abs/2204.09269, 2022.

\bibitem[Vijayakumar et~al.(2016)Vijayakumar, Cogswell, Selvaraju, Sun, Lee,
  Crandall, and Batra]{journals/corr/VijayakumarCSSL16}
Ashwin~K. Vijayakumar, Michael Cogswell, Ramprasaath~R. Selvaraju, Qing Sun,
  Stefan Lee, David~J. Crandall, and Dhruv Batra.
\newblock Diverse beam search: Decoding diverse solutions from neural sequence
  models.
\newblock \emph{CoRR}, abs/1610.02424, 2016.

\bibitem[Meister et~al.(2022)Meister, Pimentel, Wiher, and
  Cotterell]{journals/corr/abs-2202-00666}
Clara Meister, Tiago Pimentel, Gian Wiher, and Ryan Cotterell.
\newblock Typical decoding for natural language generation.
\newblock \emph{CoRR}, abs/2202.00666, 2022.

\bibitem[Fan et~al.(2018)Fan, Lewis, and Dauphin]{LewisDF18}
Angela Fan, Mike Lewis, and Yann~N. Dauphin.
\newblock Hierarchical neural story generation.
\newblock In \emph{{ACL} {(1)}}, pages 889--898. Association for Computational
  Linguistics, 2018.

\bibitem[Holtzman et~al.(2020)Holtzman, Buys, Du, Forbes, and
  Choi]{HoltzmanBDFC20}
Ari Holtzman, Jan Buys, Li~Du, Maxwell Forbes, and Yejin Choi.
\newblock The curious case of neural text degeneration.
\newblock In \emph{{ICLR}}. OpenReview.net, 2020.

\bibitem[Song et~al.(2021)Song, Meng, and Ermon]{DBLP:conf/iclr/SongME21}
Jiaming Song, Chenlin Meng, and Stefano Ermon.
\newblock Denoising diffusion implicit models.
\newblock In \emph{9th International Conference on Learning Representations,
  {ICLR} 2021, Virtual Event, Austria, May 3-7, 2021}. OpenReview.net, 2021.
\newblock URL \url{https://openreview.net/forum?id=St1giarCHLP}.

\bibitem[Dong et~al.(2023)Dong, Li, Gong, Chen, Li, Shen, and
  Yang]{DBLP:journals/csur/DongLGCLSY23}
Chenhe Dong, Yinghui Li, Haifan Gong, Miaoxin Chen, Junxin Li, Ying Shen, and
  Min Yang.
\newblock A survey of natural language generation.
\newblock \emph{{ACM} Comput. Surv.}, 55\penalty0 (8):\penalty0 173:1--173:38,
  2023.
\newblock \doi{10.1145/3554727}.
\newblock URL \url{https://doi.org/10.1145/3554727}.

\bibitem[Hoogeboom et~al.(2022)Hoogeboom, Gritsenko, Bastings, Poole, van~den
  Berg, and Salimans]{hoogeboom2022autoregressive}
Emiel Hoogeboom, Alexey~A. Gritsenko, Jasmijn Bastings, Ben Poole, Rianne
  van~den Berg, and Tim Salimans.
\newblock Autoregressive diffusion models.
\newblock In \emph{International Conference on Learning Representations}, 2022.
\newblock URL \url{https://openreview.net/forum?id=Lm8T39vLDTE}.

\bibitem[Rasul et~al.(2021)Rasul, Seward, Schuster, and
  Vollgraf]{pmlr-v139-rasul21a}
Kashif Rasul, Calvin Seward, Ingmar Schuster, and Roland Vollgraf.
\newblock Autoregressive denoising diffusion models for multivariate
  probabilistic time series forecasting.
\newblock In Marina Meila and Tong Zhang, editors, \emph{Proceedings of the
  38th International Conference on Machine Learning}, volume 139 of
  \emph{Proceedings of Machine Learning Research}, pages 8857--8868. PMLR,
  18--24 Jul 2021.
\newblock URL \url{https://proceedings.mlr.press/v139/rasul21a.html}.

\bibitem[Luo(2022)]{luo2022understanding}
Calvin Luo.
\newblock Understanding diffusion models: A unified perspective.
\newblock \emph{arXiv preprint arXiv:2208.11970}, 2022.

\end{thebibliography}

\newpage
\appendix

\section{Proof of Inference with Skipping}
\label{appendix:proof}

During the inference process, skipping strategy requires the model $\vg_{\theta}$ to infer the state $\vz^{n_2}_{t_{i+1}}$ at a far-off timestep $t_{i+1}$ compared to the current state $\vz^{n_2}_{t_{i}}$, where $t_{i+1}\ll t_i$. In our model, due to the dynamic speed setting, token $\vz^{n_1}_{t_{i+1}}$ with smaller timestep $t_{i+1}\le t_i$, which is closer to $t_{i+1}$, and positions $n_1\le n_2$ can provide stronger auxiliary information than $\vz^{n_1}_{t_i}$. This reduces the difficulty of inferring states for tokens in the end, making our multi-level diffusion model particularly suitable for accelerating the generation process.

Through maximizing the evidence lower bound (ELBO) of $p(\vz_0)$, the training object is equivalent to minimize the divergence between $q(\vz_{t}|\vz_{t-1}, \vz_0)$ and $p_{\theta}(\vz_{t-1}|\vz_{t})$ following~\citep{luo2022understanding}. 

By converting the joint probability distribution into a conditional probability distribution, we obtain the following formula for $q(\vz_{t_{i+1}}|\vz_{t_{i}}, \vz_0)$.
\begin{equation}
\small
\begin{aligned}
q(\vz_{t_{i+1}}|\vz_{t_{i}},\vz_{0})&=q(\vz_{t_{i+1}}|\vz_{t_{i}-1},\vz_{t_{i}},\vz_{0})\ q(\vz_{t_{i+1}-1}|\vz_{t_{i}},\vz_{0}) \\
&=q(\vz_{t_{i+1}}|\vz_{t_{i}-1},\vz_{0})\ q(\vz_{t_{i+1}-1}|\vz_{t_{i}},\vz_{0}) \\
&=q(\vz_{t_{i+1}}|\vz_{t_{i}-2},\vz_{0})\ q(\vz_{t_{i+1}-2}|\vz_{t_{i}-1},\vz_{0}) \ q(\vz_{t_{i+1}-1}|\vz_{t_{i}},\vz_{0}) \\
&=\prod_{k=1}^{t_{i}-t_{i+1}} q(\vz_{t_{i}-k}|\vz_{t_{i}-k+1},\vz_{0}) \label{equ:appendix:q_decomp2}
\end{aligned} 
\end{equation}
Similarly, we reach the same conclusion regarding $p_{\theta}(\vz_{t_{i+1}}|\vz_{t_{i}})$. 

Based on~\cref{equ:appendix:q_decomp2}, which consists of $q(\vz_{t}|\vz_{t-1}, \vz_0)$, and the interchangeability between $q(\vz_{t}|\vz_{t-1}, \vz_0)$ and $p_{\theta}(\vz_{t-1}|\vz_{t})$, we can decompose $q(\vz_{t_{i+1}}|\vz_{t_i}, \vz_0)$ by incorporating $\vz_{t_i}$ and $\vz_0$, and utilize our estimated $\vz_0$ to determine the expression of $p_{\theta}(\vz_{t_{i+1}}|\vz_{t_i})$.

\begin{equation}
\small
\begin{gathered}
q(\vz_{t_{i+1}}\mid \vz_{t_{i}},\vz_{0})= \prod_{n=1}^{N} q\big(z^{n}_{f(n, t_{i+1})}\mid z^{n}_{f(n, t_{i})}, z_{0}^{n}\big) 
\end{gathered} 
\end{equation}

Next, we obtain the explicit expression $q\big(z^{n}_{f(n, t_{i+1})}\mid z^{n}_{f(n, t_{i})}, z_{0}^{n}\big)$ through linear interpolation between $z^{n}_{f(n, t_{i})}$ and $z_{0}^{n}$.
\begin{equation}
\small
\begin{aligned}
  &q\big(z^{n}_{f(n, t_{i+1})}\mid z^{n}_{f(n, t_{i})}, z_{0}^{n}\big)=\frac{q(z^{n}_{f(n, t_{i})} \mid z^{n}_{f(n, t_{i+1})}, z^n_0)q(z^{n}_{f(n, t_{i+1})} \mid z^n_0)}{q(z^{n}_{f(n, t_{i})}\mid z^n_0)} \\
  =&\frac{\mathcal{N}\Big(z^{n}_{f(n, t_{i})}; \sqrt{\frac{\bar{\alpha}_{f(n, t_i)}}{\bar{\alpha}_{f(n, t_{i+1})}}}z^n_{f(n, t_{i+1})}, \big(1-\frac{\bar{\alpha}_{f(n,t_i)}}{\bar{\alpha}_{f(n,t_{i+1}})}\big)I\Big)\mathcal{N}\Big(z^{n}_{f(n, t_{i+1})}; \sqrt{\bar{\alpha}_{f(n,t_{i+1})}}z^n_0, (1-\bar{\alpha}_{f(n,t_{i+1})})I\Big)}{\mathcal{N}\Big(z^{n}_{f(n, t_{i})}; \sqrt{\bar{\alpha}_{f(n,t_i)}}z^n_0, (1-\bar{\alpha}_{t_i})I\Big)} \\
  \propto &\exp\Big\{- \frac{\big(z^{n}_{f(n, t_{i})}-\sqrt{\frac{\bar{\alpha}_{f(n,t_i)}}{\bar{\alpha}_{f(n,t_{i+1})}}}z^{n}_{f(n, t_{i+1})}\big)^2}{2(1-\frac{\bar{\alpha}_{f(n,t_i)}}{\bar{\alpha}_{f(n,t_{i+1})}})} -\frac{\big(z^{n}_{f(n, t_{i+1})}-\sqrt{\bar{\alpha}_{t_{i+1}}}z^n_0\big)^2}{1-\bar{\alpha}_{f(n,t_{i+1})}} \\
  &\quad\quad\quad + \frac{\big(z^{n}_{f(n, t_{i})}-\sqrt{\bar{\alpha}_{f(n,t_i)}}z^n_0\big)^2}{1-\bar{\alpha}_{f(n,t_i)}}\Big\} \\
  =&\exp\Big\{-\frac{1-\bar{\alpha}_{t_i}}{2(1-\frac{\bar{\alpha}_{t_i}}{\bar{\alpha}_{t_{i+1}}})(1-\bar{\alpha}_{t_{i+1}})}\Big[{z^{n}_{f(n, t_{i+1})}}^2-2\Big(\frac{\sqrt{\frac{\bar{\alpha}_{f(n,t_{i})}}{\bar{\alpha}_{f(n,t_{i+1})}}}(1-\bar{\alpha}_{f(n,t_{i+1})})}{1-\bar{\alpha}_{f(n,t_{i})}}z^{n}_{f(n, t_{i})} \\
  &\quad\quad\quad + \frac{\sqrt{\bar{\alpha}_{f(n,t_{i+1})}}(1-\frac{\bar{\alpha}_{f(n,t_{i})}}{\bar{\alpha}_{f(n,t_{i+1})}})}{1-\bar{\alpha}_{f(n,t_{i})}}z^{n}_{0} \Big)z^{n}_{f(n, t_{i+1})}\Big] \Big\} \\
  \propto&\ \mathcal{N}\Big(z^{n}_{f(n, t_{i+1})};\frac{\sqrt{\frac{\bar{\alpha}_{f(n,t_{i})}}{\bar{\alpha}_{f(n,t_{i+1})}}}(1-\bar{\alpha}_{f(n,t_{i+1})})}{1-\bar{\alpha}_{f(n,t_{i})}}z^{n}_{f(n, t_{i})}+\frac{\sqrt{\bar{\alpha}_{f(n,t_{i+1})}}(1-\frac{\bar{\alpha}_{f(n,t_{i})}}{\bar{\alpha}_{f(n,t_{i+1})}})}{1-\bar{\alpha}_{f(n,t_{i})}}z^{n}_{0}, \\
  &\quad\quad\quad  \frac{(1-\frac{\bar{\alpha}_{t_i}}{\bar{\alpha}_{t_{i+1}}})(1-\bar{\alpha}_{t_{i+1}})}{1-\bar{\alpha}_{t_i}}I\Big) \\
  =&\ \mathcal{N}(z^{n}_{f(n, t_{i+1})}; \lambda z^{n}_{f(n, t_{i})} + \mu z^{n}_{0}, \sigma I ) \label{equ:appendix:2}
\end{aligned}
\end{equation}
where we have the following notations for simplification.  
\begin{equation}
\small
\begin{aligned}
    \lambda =\frac{\sqrt{\frac{\bar{\alpha}_{f(n,t_{i})}}{\bar{\alpha}_{f(n,t_{i+1})}}}(1-\bar{\alpha}_{f(n,t_{i+1})})}{1-\bar{\alpha}_{f(n,t_{i})}},~
    \mu =\frac{\sqrt{\bar{\alpha}_{f(n,t_{i+1})}}(1-\frac{\bar{\alpha}_{f(n,t_{i})}}{\bar{\alpha}_{f(n,t_{i+1})}})}{1-\bar{\alpha}_{f(n,t_{i})}},~
    \sigma = \frac{(1-\alpha_{f(n,t_{i})})(1-\bar{\alpha}_{f(n,t_{i+1})})}{1-\bar{\alpha}_{f(n,t_{i})}} \nonumber
\end{aligned}
\label{equ:appendix:3}
\end{equation}

Building upon~\cref{equ:appendix:2}, we substitute $\vz^n_{0}$ with $\vg_{\theta}(\vz^n_{f(n,t)}, f(n,t);\vx)$, yielding the final formula for $p_{\theta}\big(\vz^{n}_{f(n, t_{i+1})}\mid \vz^{n}_{f(n, t_{i})};\vx\big)$ as the following equation.
\begin{equation}
\small
\begin{gathered}
p_{\theta}\big(\vz^{n}_{f(n, t_{i+1})}\mid \vz^{n}_{f(n, t_{i})};\vx\big) \sim \mathcal{N}\big(\vz^{n}_{f(n, t_{i+1})};\lambda \vz_{f(n,t_{i})}^{n}+\mu\vg_{\theta}(\vz^n_{f(n,t)}, f(n,t);\vx), \sigma \textbf{I}\big) 
\end{gathered}
\label{equ:appendix:reverse_sample2_3}
\end{equation}

\section{More Cases}
\label{appendix:cases}






\begin{figure}[ht]
	\centering
	\begin{minipage}[ht]{\linewidth}
		\centering
		\includegraphics[width=\linewidth]{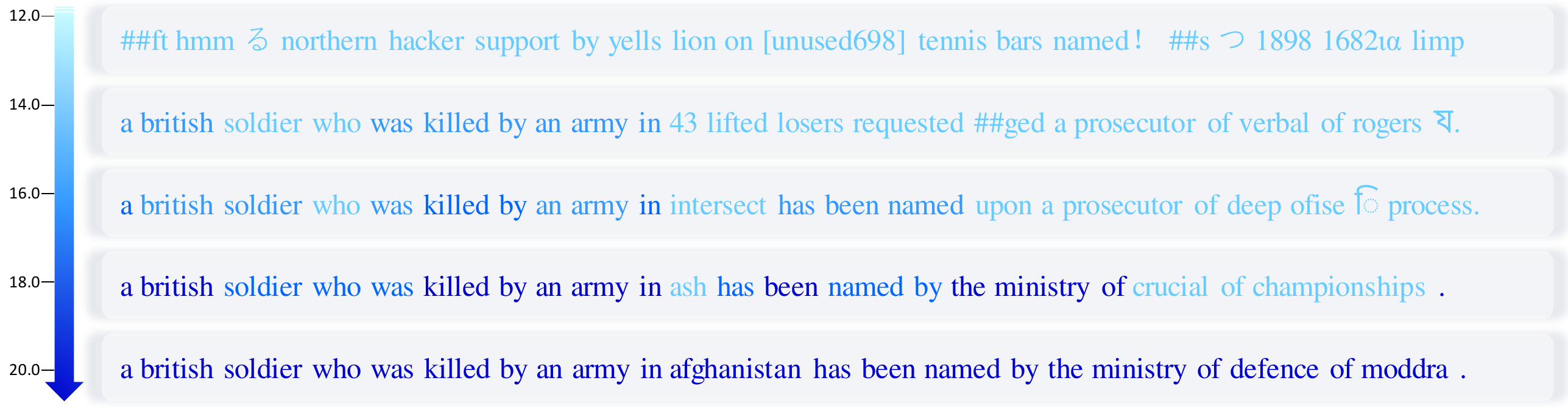}
	\end{minipage}
	\\
	\begin{minipage}[ht]{\linewidth}
		\centering
		\includegraphics[width=\linewidth]{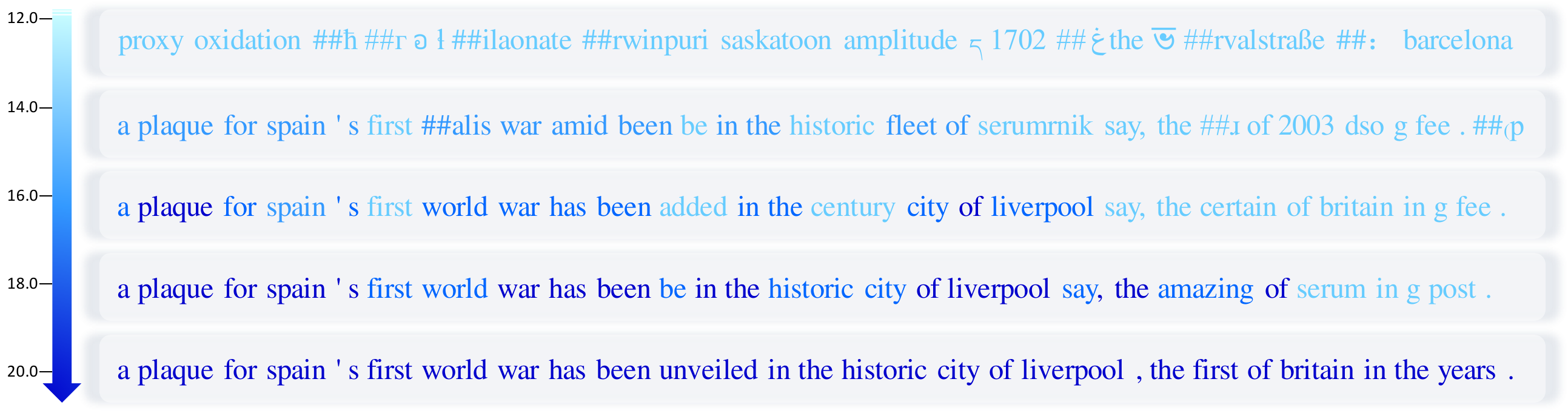}
	\end{minipage}
 \\
	\begin{minipage}[ht]{\linewidth}
		\centering
		\includegraphics[width=\linewidth]{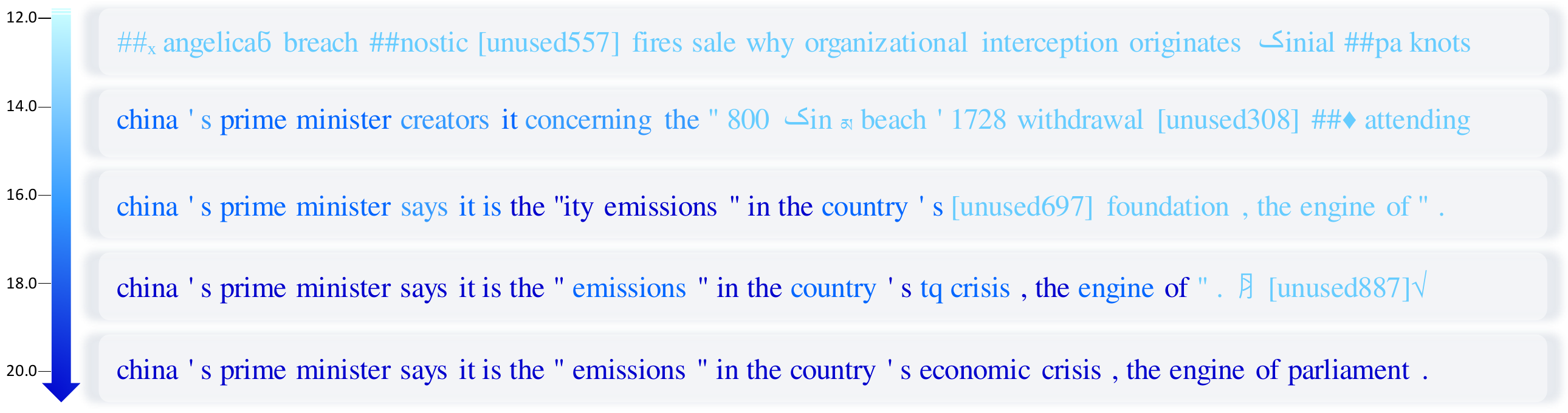}
	\end{minipage}
\end{figure}

\end{document}